\documentclass[journal,onecolumn]{IEEEtran}
\usepackage{graphicx}
\usepackage{cite}
\usepackage{amsmath,amssymb,amsfonts}
\usepackage{algorithmic}
\usepackage{textcomp}
\usepackage{xcolor}
\usepackage{subcaption}
\usepackage{multirow}
\usepackage{gensymb}
\usepackage{color,soul}
\usepackage{float}
\usepackage{hyperref}
\usepackage{fancyhdr} 
\usepackage{textcomp} 
\usepackage{array}    

\makeatletter
\newcolumntype{"}{@{\tabcolsep\vrule width 1pt\hskip\tabcolsep}}
\newcolumntype{'}{@{\tabcolsep\hskip\vrule width 1pt\tabcolsep}}
\makeatother

\ifCLASSINFOpdf

\else

\fi

\begin{document}

\title{PaXNet: Dental Caries Detection in Panoramic X-ray using Ensemble Transfer Learning and Capsule Classifier}

\author{
    \IEEEauthorblockN{Arman Haghanifar\IEEEauthorrefmark{1}\IEEEauthorrefmark{3}, Mahdiyar Molahasani Majdabadi\IEEEauthorrefmark{2}\IEEEauthorrefmark{3}, Seok-Bum Ko\IEEEauthorrefmark{1}\IEEEauthorrefmark{2}
}
    
    \IEEEauthorblockA{\IEEEauthorrefmark{1} Div. of Biomedical Engineering, University of Saskatchewan}
    
    \IEEEauthorblockA{\IEEEauthorrefmark{2} Dept. of Electrical and Computer Engineering, University of Saskatchewan}
    
    \IEEEauthorblockA{\IEEEauthorrefmark{3} Both authors contributed equally to this manuscript }
}

\maketitle

\begin{abstract}
Dental caries is one of the most chronic diseases involving the majority of the population during their lifetime. Caries lesions are typically diagnosed by radiologists relying only on their visual inspection to detect via dental x-rays. In many cases, dental caries is hard to identify using x-rays and can be misinterpreted as shadows due to different reasons such as low image quality. Hence, developing a decision support system for caries detection has been a topic of interest in recent years. Here, we propose an automatic diagnosis system to detect dental caries in Panoramic images for the first time, to the best of authors' knowledge. The proposed model benefits from various pretrained deep learning models through transfer learning to extract relevant features from x-rays and uses a capsule network to draw prediction results. On a dataset of 470 Panoramic images used for features extraction, including 240 labeled images for classification, our model achieved an accuracy score of 86.05\% on the test set. The obtained score demonstrates acceptable detection performance and an increase in caries detection speed, as long as the challenges of using Panoramic x-rays of real patients are taken into account. Among images with caries lesions in the test set, our model acquired recall scores of 69.44\% and 90.52\% for mild and severe ones, confirming the fact that severe caries spots are more straightforward to detect and efficient mild caries detection needs a more robust and larger dataset. Considering the novelty of current research study as using Panoramic images, this work is a step towards developing a fully automated efficient decision support system to assist domain experts.
\end{abstract}



\thispagestyle{fancyplain}
\lfoot{\textit{This work is the extended version of "Automated Teeth Extraction from Dental Panoramic X-Ray Images using Genetic Algorithm," \cite{haghanifar2020automated} in 2020 IEEE International Symposium on Circuits and Systems (ISCAS), Seville, Spain, Oct. 2020.}}

\section{Introduction}\label{intro}
Dental caries, also known as tooth decay, is one of the most prevalent infectious chronic dental diseases in humans, affecting individuals throughout their lifetime \cite{selwitz2007dental}. According to the National Health and Nutrition Examination Survey, dental caries involves approximately 90\% of adults in the United States \cite{amrollahi2016recent}, \cite{beltran2005surveillance}. Dental caries is a dynamic disease procedure resulting from dental biofilm's metabolic activity, which gradually demineralizes enamel and dentine \cite{pitts2016white}. Tooth decay is a preventable disease, and if detected, can be stopped and potentially reversed in its early stages \cite{fejerskov2009dental}. A standard tool for radiologists to distinguish dental diseases, such as caries, is x-ray radiography. X-rays are an essential complementary diagnosis solution to help identifying dental problems that are hard to detect via visual inspection only \cite{lira2009panoramic}.

Radiography is one of the methods usually performed to help radiologists with the oral health assessment and diagnosis of dental diseases, such as proximal dental caries \cite{tagliaferro2019caries}. Since proximal tooth surfaces are hard to be approached or visualized directly, caries lesions in these surfaces are diagnosed with the aid of radiographs \cite{qu2011detection}. There are several dental x-ray types, each of which records a different view of dental anatomies, such as Bitewing, Panoramic, and Periapical. Basically, Bitewing radiography is the most widely used method for caries detection and has the highest diagnostic accuracy, while Panoramic has the lowest \cite{akarslan2008comparison}. However, Bitewing has several disadvantages, like patient discomfort. Due to operators' insufficient expertise, Bitewing imaging usually results in increased patient radiation dose because of the need for image retakes \cite{casamassimo1981radiographic}. On the other hand, Panoramic x-ray images, also known as Orthopantomogram (OPG), are widely used to capture the entire mouth using a very small dose of radiation \cite{underhill1988radiobiologic}. OPG has a low radiation dose, simplicity of application, less time requirement, and also great patient comfort. Thus, pediatric, handicapped, and senior patients would benefit greatly from a Panoramic imaging system compared to intraoral systems \cite{akkaya2006comparing}.

Since a Panoramic image covers the entire patient dentition along with surrounding bones and jaw structure, it can not give a detailed view of each tooth. Hence, structures in Panoramic images lack specific boundaries, and visual quality is extremely low in comparison with other types of dental radiographs \cite{naam2016algorithm}. Besides, Panoramic images also include other parts of the mouth, such as jawbones, which make the image analysis difficult \cite{jader2018deep}. Hence, image preprocessing steps and teeth extraction are needed to facilitate visual interpretations and enhance model performance on dental disease detection \cite{naam2016algorithm}. A limited number of research studies have been conducted on extracting single tooth from Panoramic images using various methods, such as deep learning \cite{lee2020application} or evolutionary algorithms \cite{haghanifar2020automated}. The latter mentioned research is introduced as the first step toward creating a fully automated decision support system. In this work, the teeth extraction model is expanded to an end-to-end caries diagnosis model. This system has been utilized in order to provide the required data for training the proposed classifier in this paper. Bringing all together, Panoramic imaging is an affordable method accessible to most patients, covering a large maxillofacial segment including all the teeth. However, these images are noisy, low-resolution, and also need further preprocessing steps.

Recently, a number of research studies have proposed deep learning-based Computer-Aided Diagnosis (CAD) systems to detect dental caries based on various types of data, including clinical assessments \cite{haghanifar2018dental}, infrared light \cite{fried2020detecting}, or near-infrared transillumination imaging \cite{casalegno2019caries}. Since x-ray radiography is the most common imaging modality in dental clinical practice, the majority of studies have utilized x-rays to develop decision support systems for tooth decay diagnosis. Srivastava \textit{et al.} developed a deep fully Convolutional Neural Network (CNN)-based CAD system and applied their model on a large dataset of 3000 Bitewing x-rays to detect dental caries, which outperformed certified dentists in terms of overall f1-score \cite{srivastava2017detection}. Regarding Periapical x-rays, there have been some works in recent years. In 2016, Choi \textit{et al.} trained a model based on simple CNN architecture along with crown extraction algorithm using 475 Periapical images to boost the detection rate of proximal dental caries \cite{choi2016boosting}. Later in 2018, Lee \textit{et al.} used a pretrained Inception V3 for transfer learning on a set of 3000 Periapical images to diagnose dental caries \cite{lee2018detection}. In the most recent research study, Khan \textit{et al.} benefited from a specialist-labeled dataset of 206 Periapical radiographs and trained a U-Net to segment three different dental abnormalities, such as caries \cite{khan2020automated}. While there have been some works utilizing OPG images, such as classification of tooth types with a Faster-Regional Convolutional Neural Network (Faster-RCNN) \cite{laishram2020detection}, to the best of authors' knowledge, there are no studies applying deep neural networks on Panoramic images to detect dental caries.

Teeth extraction is an essential part needed for developing an automatic dental caries detection system that helps increasing model accuracy by preparing extracted tooth images as model input. Teeth extraction or isolation is the process of extracting image parts, each containing one tooth's boundaries, from a dental x-ray image that also contains other unwanted parts of the mouth, like gingivae or jawbones. Automatic teeth extraction module eliminates the need for manual annotation of teeth in Panoramic images needed for both developing dental disease detection systems or training deep learning-based teeth segmentation models. In \cite{haghanifar2020automated}, we have proposed a novel genetic-based approach for teeth extraction in Panoramic dental images. Based on a dataset of Panoramic x-rays, our teeth extractor could isolate teeth with accuracy scores in line with previous works utilizing either Bitewing or Periapical x-rays. Various methods were introduced for jaw separation, teeth isolation, and accuracy improvement of the system. Considering the aforementioned teeth extraction system as the preliminary step to prepare single tooth images for the disease classification model, current study aims to extend the previous system to build an end-to-end caries detection system where a digital OPG image is the input and teeth suspicious of having caries are the final output.

Since there have been limited attempts to develop automatic deep learning-based dental disease diagnosis systems, there is a need to perform further research in this area. The objective of this study is to develop a specialized model architecture based on pretrained models and the capsule network to detect tooth decay on Panoramic x-rays efficiently. To the best of authors' knowledge,

\begin{itemize}
    \item This research is the first to perform teeth extraction as well as dental caries detection on Panoramic images, using a relatively large dataset. Most previous works addressed other types of dental x-rays with higher quality in terms of noise level and resolution.  
    \item Genetic algorithm is applied for the first time to isolate teeth in Panoramic images. Previous studies rely mainly on manually defined methods. This evolutionary algorithm demonstrates robust performance even on challenging jaws with several missing teeth. 
    \item Capsule network is used for the first time as the classifier for dental caries diagnosis. Experimental results demonstrate its superiority over CNNs because of the fact that the Capsule network is capable of learning the geometrical relationships between features.
    \item Feature extraction module is constructed by a voting system from different pretrained architectures. CheXNet \cite{rajpurkar2017chexnet} is applied for the first time in dental disease detection tasks using x-rays.
\end{itemize}

The paper is structured as follows. After the introduction, section II is about the materials used for this study and the labeling process required. Section III explains the model architecture both for feature extraction and classification parts. Results are presented and thoroughly discussed in section IV. Finally, section V is the conclusion.

\section{Materials and Dataset Preparation}
Most related studies in the field of dental problem detection using x-rays lack a sufficient number of images in their datasets. Large datasets let the models have more sophisticated architectures, including more parameters. Hence, developed models can handle more complicated features and detect subtle abnormalities that appeared in the tooth texture, such as dental caries in the early stages. Annotation is an essential and time-consuming part that needs to be performed by the field specialists, e.g. dentists or radiologists.

\subsection{Dataset Collection}
\begin{figure}[H]
    \centering
    \begin{subfigure}{0.45\linewidth}
        \centering
        \includegraphics[width=0.9\linewidth]{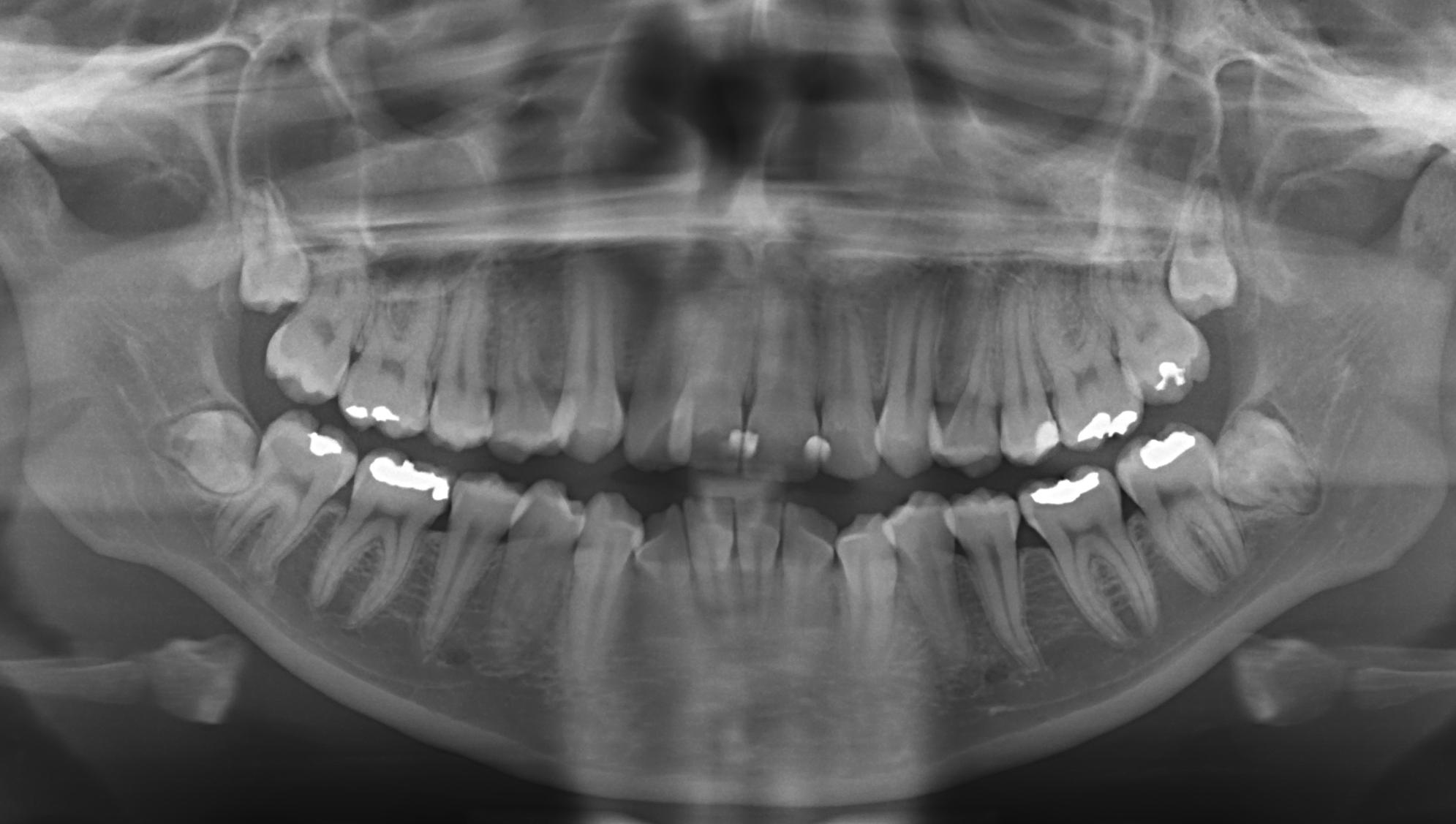}
        \caption{}
        \label{fig:test_uesb}
    \end{subfigure}
    \begin{subfigure}{0.45\linewidth}
        \centering
        \includegraphics[width=0.9\linewidth]{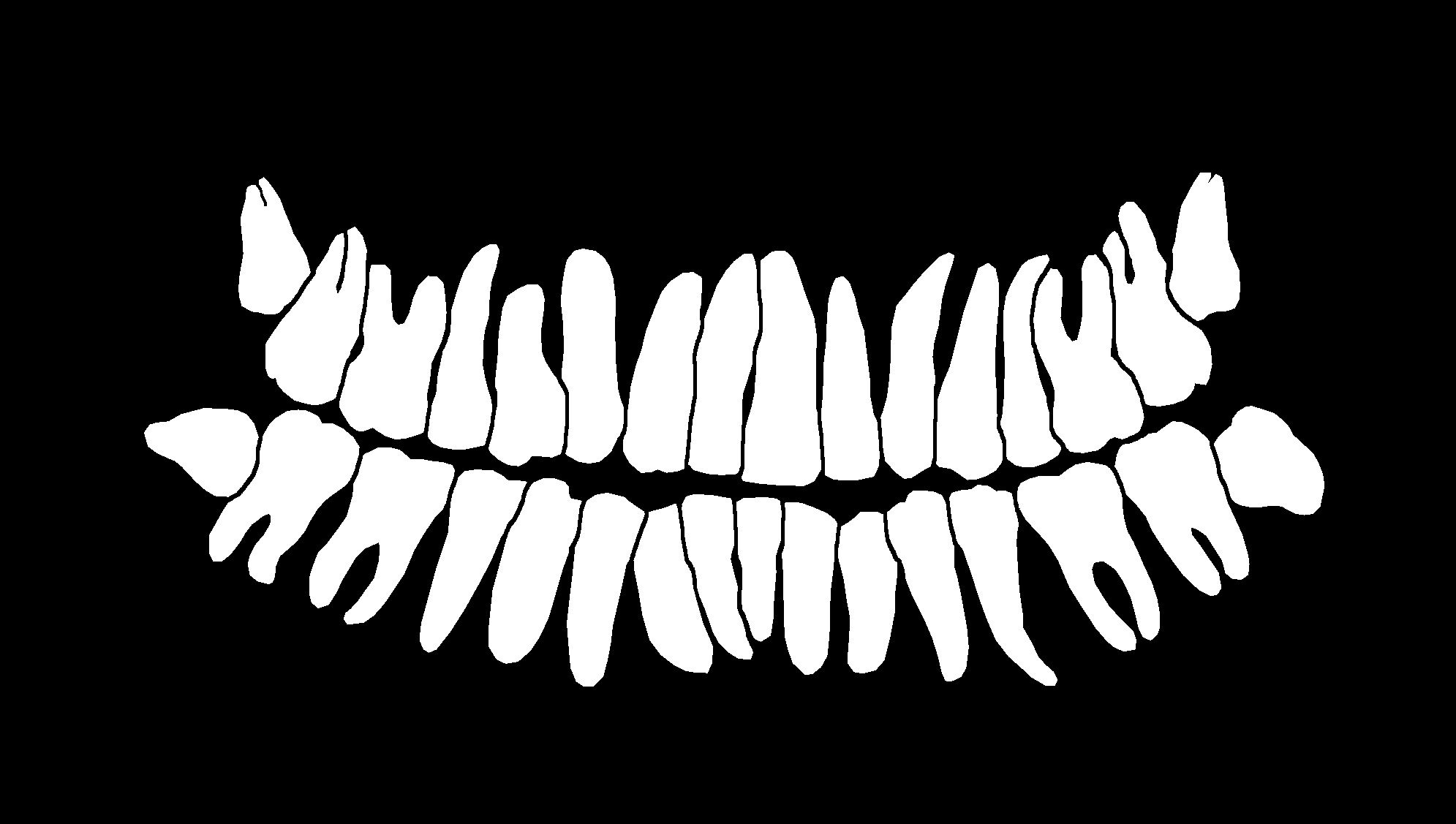}
        \caption{}
        \label{fig:test_uesb_mask}
    \end{subfigure}
\caption{(a) An example image from the UESB dataset with (b) the related tooth annotation mask \cite{silva2018automatic}}
\label{test_uesb}
\end{figure}

Our dataset of $470$ Panoramic x-rays is collected from two main sources along with a few number of images from publicly available resources. $280$ images are obtained from the Diagnostic Imaging Center of the Southwest State University of Bahia (UESB) \cite{silva2018automatic}. Images are acquired from the x-ray camera model ORTHOPHOS XG 5 DS/Ceph from Sirona Dental Systems GmbH. Images are randomly selected from different categories with an initial size of $1991 \times 1127$. All images are in "JPG" format. Annotation masks related to the images are available in the UESB dataset. An example OPG image from the dataset is shown in Fig. \ref{test_uesb}.

Besides, we also collected $120$ images from a local dentistry clinic taken with x-ray camera model Cranex\textregistered 3D from Soredex. Images are anonymized and obtained with a contrast pre-enhancement applied by the radiologist. All images are in "BMP" format with an initial size of $3292 \times 1536$ with the bit-depth of $8$. $42$ images from our dataset are randomly selected to validate the performance of the teeth extraction system. $70$ images are downloaded from publicly available medical image sources, such as Radiopaedia \footnote{\href{https://radiopaedia.org}{https://radiopaedia.org}}. Public images are taken with unknown camera models and are available in different sizes and formats. Some of the collected images are associated with a radiologist report indicating locations of tooth decays. The dataset includes $11769$ images from single tooth, which is used for training the encoder model. Among the dataset, $240$ images have labeled teeth, including $742$ carious and $5206$ healthy.

\subsection{Labeling Process}
UESB images have tooth masks manually prepared along with the dataset, which is used to extract each tooth from Panoramic images. For our collected dataset, a genetic-based method is applied to isolate the teeth by finding the optimum lines which fall inside gaps between teeth in both maxillary, and mandibular jaws \cite{haghanifar2020automated}. To perform labeling, a radiologist commented on Panoramic images one by one. Extracted teeth are categorized into two groups; healthy and carious. Carious teeth are also classified into mild and severe caries. Mild ones are caries lesions in their early stages, mostly located in enamel or Dentine-Enamel Junction (DEJ). In contrast, severe decays are developed dental plaques that have been spread to the internal dentine or have involved the pulpitis. Pulpitis caries lesions result in a collapsed tooth. 

The labeling process is a time-consuming and challenging task that requires huge amount of time. Since Panoramic images include all teeth in one image, it helps radiologists meticulously detect caries by not only inspecting opaque areas on the tooth but also considering the type of the tooth and its location in the jaw. On the other hand, higher levels of noise and shadows make the visual diagnosis more challenging. Caries, especially mild ones, can easily be misinterpreted as shadows and vice versa. Another problem is the Mach effect. Mach effect or Mach bands is an optical phenomenon that makes the edges of darker objects next to lighter ones appear lighter and vice versa. Mach effect results in a false shadow that may bring diagnostic misinterpretation with dental caries present very close to dental restoration regions that are appeared to be whiter in dental x-rays \cite{martinez2011radiopacity}.

\section{Model Architecture}
In this section, firstly Genetic Algorithm (GA) is introduced, and its usage in the field of image processing is briefly discussed. A teeth extraction system is then presented, and details of different modules are investigated one by one. Afterwards, the architecture and advantages of using capsule network are reviewed. Finally, the feature extraction unit's architecture and the classifier are explained, followed by a detailed illustration of the proposed PaXNet.

\subsection{Genetic Algorithm}\label{GA}
A genetic algorithm is considered a blind optimization technique that mimics natural evolution's learning process using selection, crossover, and mutation. These three procedures are transformed into numerical functions to help solve an optimization problem without calculating derivatives. Using a random exploration of the search space, GA is more robust in terms of being stuck in the local extrema \cite{goldberg2006genetic}. In image processing, GA is proven as a powerful search method that converts an image segmentation problem into an optimization problem \cite{sheta2012genetic}.

In this research study, teeth are isolated using separator lines that pass the gap between two teeth next to each other. To perform the task, GA is proposed to explore the solution space, in the sense that separator lines fit into the paths with the lowest integral intensity. These paths are considered as including gaps between proximal teeth surfaces. The GA can dynamically change the image segmentation task controlling parameters to reach the best performance. The GA-based teeth extraction process is summarized as follows:

\begin{itemize}

    \item Initial population: A number of random lines with a limited degree of freedom are considered with a certain distance from each other
    
    \item Cost function: The integral intensity projection of all the pixels in each line is considered as the cost function which needs to be minimized; meaning that the lines pass through the darkest available path in the area of proximal surfaces of two neighbor teeth. Cost function is formulated as follows:
    
    \begin{equation} \label{cost_fn}
        C(x)= \sum_{i=1}^{n} I(x_i)
    \end{equation}
    
    \noindent where $x$ is the position of each line, $n$ is the number of lines, and $I(x_i)$ is the average of the intensity of the pixels on the line $x_i$.
    
    \item Genetic cycle: Produced lines are changed during iterations and are ranked based on the above-mentioned cost function. Crossover and mutation functions are specified as Scattered and Gaussian, respectively. This cycle is performed iteratively until the maximum fitness or one of the termination criteria is reached. The genetic cycle is shown in Fig. \ref{fig:gen_cycle}.

    \begin{figure}[H]
        \centering
        \includegraphics[width=1.0\linewidth]{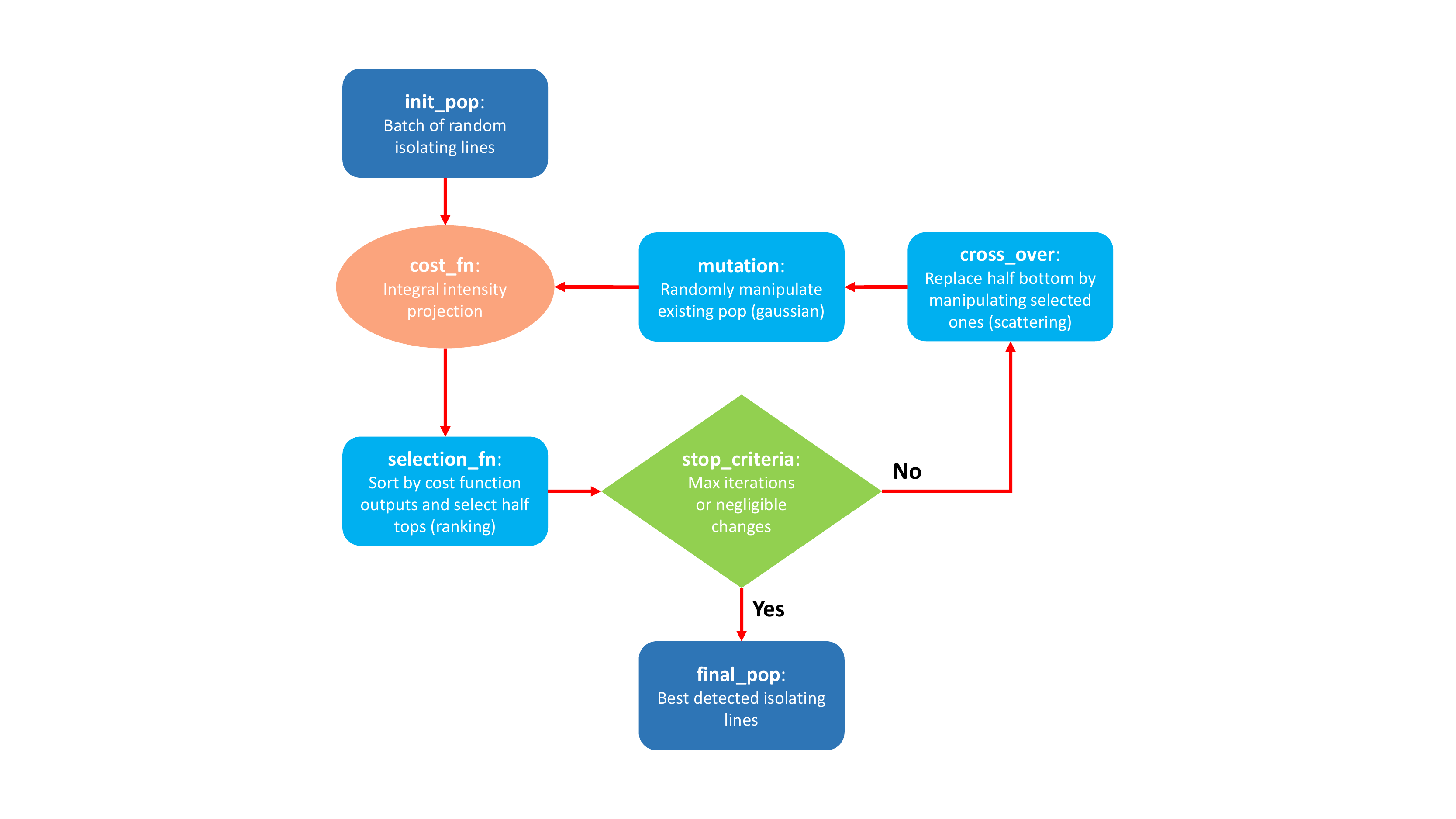}
        \caption{Flow chart of the proposed GA-based teeth isolation method}
        \label{fig:gen_cycle}
    \end{figure}
    
\end{itemize}

\subsection{Teeth Extraction}
\begin{figure}[H]
    \centering
    \includegraphics[width=0.9\linewidth]{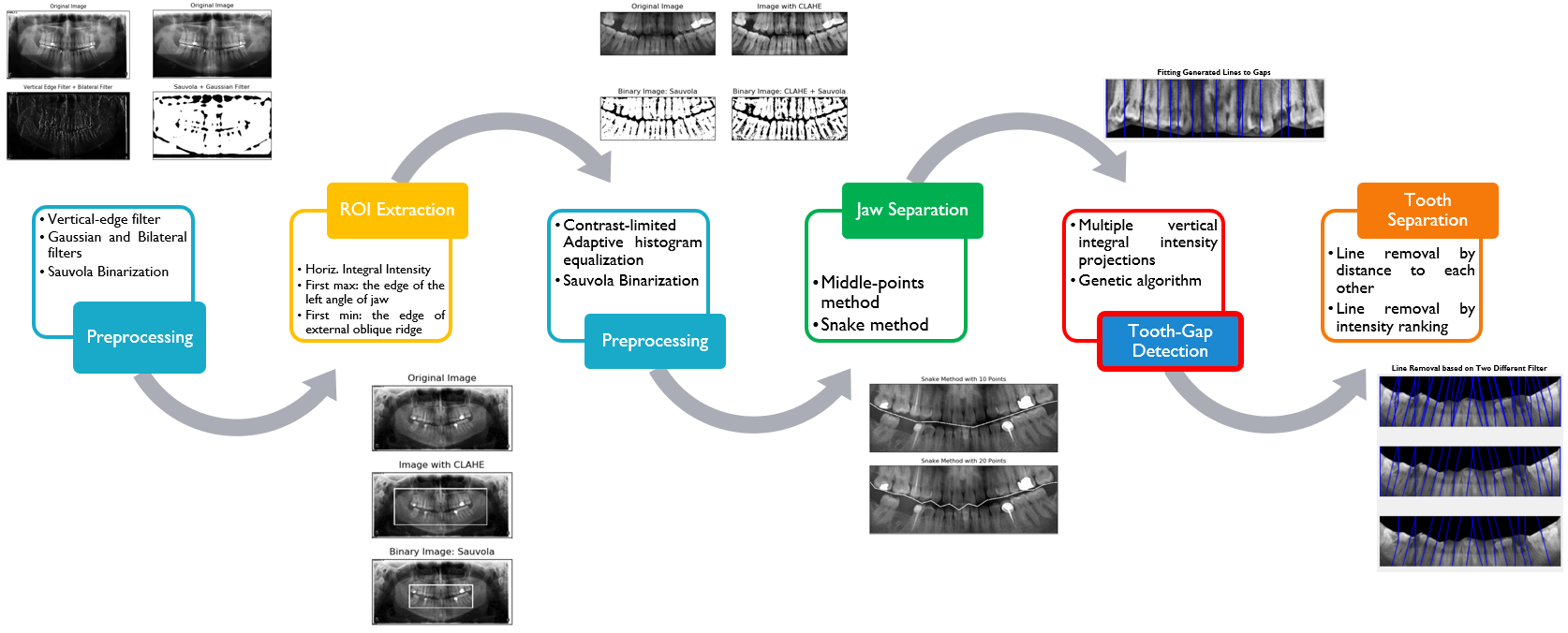}
    \caption{Detailed diagram of the proposed teeth extraction system}
    \label{fig:detailed_diagram}
\end{figure}

Since Panoramic images contain unwanted regions around the jaw, some segmentation algorithms should be applied to the image before performing jaw separation. Unlike  Periapical images, in Panoramic images, upper jaws (maxilla) and lower jaws (mandible) need to be separated apart before tooth isolation. Each step requires its own preprocessing method to help increase performance efficiency. A detailed high-level illustration of the proposed teeth extraction pipeline is depicted in Fig. \ref{fig:detailed_diagram}.

As shown in Fig. \ref{fig:detailed_diagram}, Panoramic raw images are the input of the system. Preprocessing, ROI extraction, and jaw separation are done, one after another, to make the image ready for the genetic algorithm to generate vertical lines for teeth separation. Line removal is then applied to output images, each including one tooth. These generated images are then fed to the PaXNet to detect dental caries and their severity. A brief description of teeth extraction methodology is described in the following subsections.

\subsubsection{Preprocessing}
To enhance the image quality and highlight useful details, preprocessing steps have been taken. For extracting the jaw as the Region of Interest (ROI) within the main image, two steps are considered. To extract initial ROI, a vertical-edge-detecting filter is applied to highlight the vertical edges. Then, a bilateral filter is used for sharpening the edges. After cropping the initial ROI, the Sauvola algorithm is applied to binarize the image in order to blacken the two ends of the gap between maxilla and mandible. Finally, a Gaussian filter is used to reduce the noise.

\subsubsection{ROI Extraction}
To reach the jaw from the main image, unwanted surroundings need to be discarded. ROI extraction is performed in two steps. After the preprocessing for the first step described above, horizontal integral intensity projection is applied to the image. The first significant positive slope represents the edge of the left angle of the jaw. On the next step, horizontal integral intensity projection is performed on the cropped image to distinguish the external oblique ridge, the starting point for the gap between maxilla and mandible, which appears to be the first significant negative slope in the intensity graph.

\subsubsection{Jaw Separation}
After extracting the jaw from the main image, maxilla and mandible are expected to become separate using a line that passes through the gap while maintaining the biggest possible distance from both jaws. Two procedures are employed to create the separating line: middle points and snake. In the middle points method, the image is divided into several parts. Then, vertical integral intensity is computed, and the minimum value is supposed to be the point representing the gap. The final line is eventually formed by connecting these points. Another method is the snake algorithm. First, a starting point is determined. Afterwards, it crawls through both left and right directions, looking for paths with minimum integral intensity. Each path-step length is a controlling parameter to restrict the snake from getting trapped in teeth-gap valleys.

\subsubsection{Tooth Isolation}
The last step is separating maxilla and mandible into a batch of isolated teeth. The process is similar to the jaw separation, whereas here, we have multiple lines. To find all the lines simultaneously and without any predefined parameters, a GA-based method is employed. Tooth morphology varies among the dentition, and the genetic algorithm can detect the best fitting lines due to its randomness. At first, $30$ vertical lines are randomly generated over each jaw's image as the chromosomes of the initial population. The genetic cycle discussed earlier in subsection \ref{GA} is then performed to find the best population indicating lines fitting inside gaps between teeth. Since generated lines are more than available gaps, various line removal methods are implemented afterwards to reach the number of lines precisely equal to the number of gaps in each jaw.

\subsection{Capsule Network}
\begin{figure}[H]
    \centerline{\includegraphics[width=0.6\linewidth ]{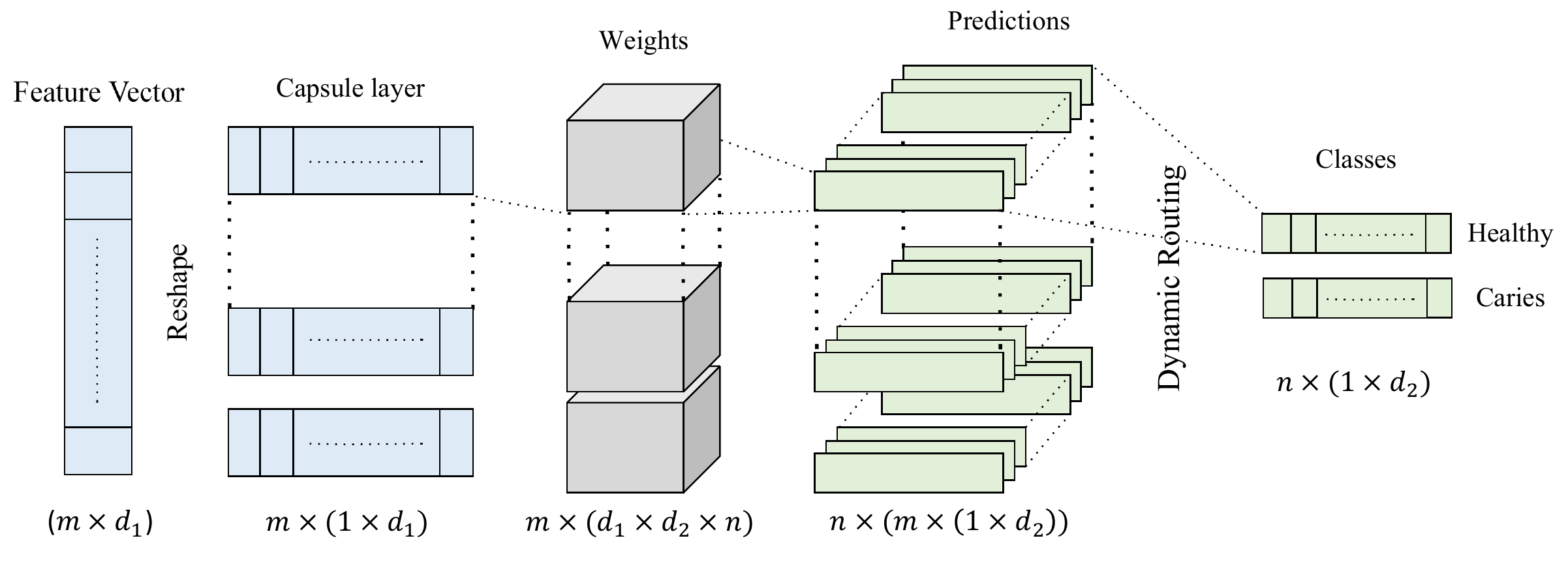}}
    \caption{The architecture of the CapsNet for binary caries classification from extracted features}
    \label{Caps}
\end{figure}

Since the introduction of capsule network \cite{NIPS2017_6975}, many studies have benefited from its advantages in various applied deep learning tasks \cite{majdabadi2020msg,pal2018capsdemm,tang2019capsurv,iesmantas2018convolutional}. Unlike convolutional networks, capsule layers represent each feature using a vector in the way that the length of the vector corresponds to the probability of the presence of a certain feature or class. A weight matrix is multiplied to each vector to predict the probability and the corresponding pose of the next level feature in the form of a multidimensional vector. Then, an algorithm called dynamic routing is applied to all the predictions of one class to determine the coherency of the predictions. This algorithm calculates a weighted average of predictions and reduces the impact of those vectors incoherent with others, iteratively. Since the average vector's length represents the probability of the class, it should be ranged between 0 and 1. In order to make sure of that, the Squash function is applied to the prediction vector after each iteration of dynamic routing, as follows:

\begin{equation} \label{capsule_loss}
    V_j=\frac{\left \| S_j \right \|^2}{1+\left \| S_j \right \|^2}\frac{S_j}{\left \| S_j \right \|}
\end{equation}

where, $S_j$ is the vector after dynamic routing. Fig \ref{Caps} indicates the structure of the tow-layer Capsule network utilized for caries detection.

\noindent Where $m$ is the number of capsules in the first layer, $d_1$ is the size of each capsule in the first layer, $n$ is the number of capsules in the number of classes (2 in this case), and $d_2$ is the size of each capsule in the second layer. The feature vector is the input of the CapsNet. This feature vector is extracted from feature extraction unit, which will be reviewed in detail in \ref{F_E}. The output of the CapsNet is two vectors representing two classes, healthy and caries. The length of these two vectors corresponds to the probability of each class. 

Moreover, this network is capable of learning the geometrical relationship between features. Hence, it can offer unprecedented robustness in a variety of tasks \cite{majdabadi2020capsule,zhao2019deep,frosst2018darccc}. The main challenge in dental caries detection in Panoramic images is distinguishing between dark parts caused by shadows and caries accurately. Geometrical features such as edges, textures and localization of these darker regions are essential for correct classification. This is why we believe the capsule network is the perfect choice for this task.

\subsection{Proposed Panoramic Dental X-ray Network: PaXNet}
The proposed network for caries detection in Panoramic dental x-rays (PaXNet) consists of two main modules; feature extraction and classifier. The aforementioned modules are explained in detail in the following subsections.

\subsubsection{Feature Extraction}\label{F_E}
As far as feature extraction is concerned, to overcome the problem of the small number of samples in the dataset, the transfer learning paradigm is used in the proposed architecture. Three pretrained models are utilized in the feature extraction block: Encoder, CheXNet, and InceptionNet. All of these three models are non-trainable, and their top layers are excluded. InceptionNet is a powerful model trained on the ImageNet dataset \cite{szegedy2015going}. Since caries appears in different sizes, this model's ability to learn features with multiple sizes is beneficial for caries detection. However, the model is trained on RGB images that share no similarities with Panoramic dental x-rays. This is why CheXNet is included in feature extraction block as well. CheXNet is a robust model for lung disease detection based on chest x-ray images \cite{rajpurkar2017chexnet}. This model is trained on CXR-14, the largest publicly available dataset of chest x-rays from adult cases with 14 different diseases. Although Panoramic dental x-rays are different from chest x-rays, since they are both x-rays, they share many similar features. There might be some features specific to a tooth that is not covered by the two models mentioned above. An encoder is used to benefit from these types of features.

The number of labeled tooth images is limited and expanding the dataset requires time and effort of specialists. In contrast, there is a considerable number of unlabeled images available. To benefit from this vast dataset, a new approach is proposed. An unsupervised side-task is designed, and a model is trained using an unlabeled dataset. Then, trained weights are used in the main model for the caries detection task. Through this approach, transfer learning enables us to take advantage of the unlabeled data as well. An auto-encoder is developed and trained to encode the image and reconstruct it from the coded version in this work.

The auto-encoder consists of two networks, Encoder and Decoder. Both of these models are used in PaXNet through transfer learning. Since all the image information should be preserved through the encoding process, an encoder can learn the most informative tooth images' features. Later, this model is used as a pretrained network for feature extraction in PaXNet. The model is benefiting from the decoder as well as the encoder, as it is explained in the next subsection.

Finally, since all these three models are non-trainable, a trainable convolutional feature extractor is embedded in PaXNet so that the model could learn task-related specific features as well. Table \ref{FE} presents a comparison between these four feature extraction units.

\begin{table}[H]
    \caption{Comparison between feature extractor models}
    \label{FE}
    \centering
    \begin{tabular}{|c|c|c|c|c|c|}
    \hline
    \textbf{Model} & \textbf{Number of Layers} & \textbf{Number of Parameters} & \textbf{Trainable} & \textbf{Dataset}          & \textbf{Dataset size} \\ \hline
    InceptionNet   & 42                        & 451,160                       & non-trainable      & ImageNet                  & 1,281,167             \\ \hline
    CheXNet        & 140                       & 1,444,928                     & non-trainable      & CXR-14                    & 112,000               \\ \hline
    Encoder        & 4                         & 14,085                        & non-trainable      & Unlabeled extracted teeth & 11,769                \\ \hline
    CNN            & 8                         & 35,808                        & trainable          & -                         & -                     \\ \hline
    \end{tabular}
\end{table}

\noindent As the similarity of the training dataset to the cries section dataset increases, the number of available samples is receded. However, more informative features can be obtained from these networks trained with more similar samples.  

As far as the activation function is concerned, all convolutional layers benefit from the Swish activation function. This function is a 
continuous replacement of leaky-ReLU, which improves the performance of the network \cite{ramachandran2017searching}. The swish activation function is formulated as follows:

\begin{equation} \label{swish_loss}
f(x) = \frac{x}{1+e^{-x}}
\end{equation}

This activation function is basically the multiplication of the Sigmoid function with the input value. The behaviour of this activation function is similar to the ReLU and leaky-ReLU in positive values. However, in large negative values, its output is converging to zero, unlike Leaky-ReLU.

\subsubsection{Classifier}
In PaXNet, all extracted features are concatenated, and higher-lever features are created based on them using a CNN. Then, the last convolutions layer is flattened, followed by a fully-connected layer with 180 neurons. These 180 neurons are reshaped to 10 capsules with eight dimensions called the primary capsule layer. There are two 32 dimensional capsules in the second capsule layer representing two classes, caries and healthy. Each capsule in the primary capsule layer makes a prediction for each capsule in the second layer. Routing by agreement is performed on these predictions for three iterations. Each vector's length is then computed, and a softmax function is applied to these two values. The output of the softmax layer is the probability of each class.

Furthermore, the capsule corresponds to the class with a higher probability is extracted using a mask. This 32D value is passed to a CNN followed by the decoder. The decoder is extracted from the auto-encoder explained in subsection \ref{F_E}. It is suggested by \cite{NIPS2017_6975} that image reconstruction can improve the capsule network's performance. Since the dataset is relatively small, the network is not able to learn the proper image reconstruction. So the decoder is utilized, and CNN is responsible for mapping the latent space of the capsule network to the decoder's latent space. 

\begin{figure}[H]
    \centerline{\includegraphics[width=0.9\linewidth ]{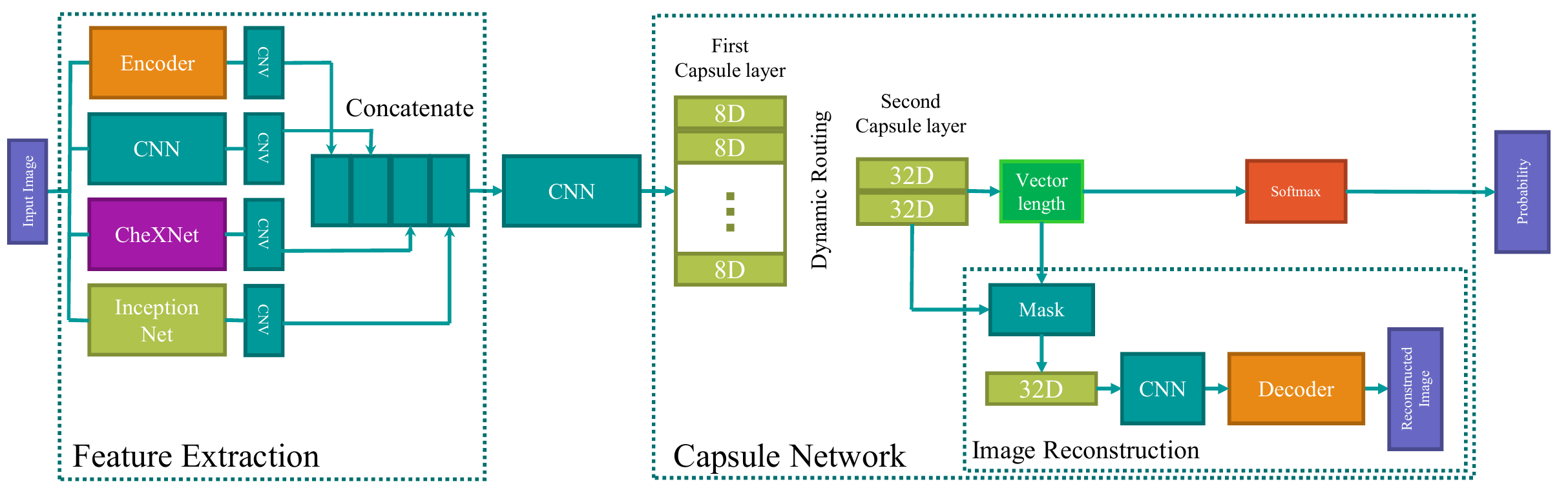}}
    \caption{Architecture of the proposed PaXNet, consisting of four networks in feature extraction module and a 2-layered capsule network to generate probability, connected with a CNN.}
    \label{Arch}
\end{figure}

The aforementioned feature extractor and classifier modules are assembled together to form PaXNet. The high-level architecture of the proposed model is illustrated in Fig. \ref{Arch}.

\noindent First, informative and valuable features of the input image are located using the feature extraction unit. Then, all these features are combined to form higher level and more complex features by a CNN. Finally, CapsNet classifies the input into two healthy or caries classes based on the extracted features. The Image reconstruction module is implemented for the training purpose. The difference between the reconstructed image and the input is used as a loss function in order to force the model to learn better features \cite{NIPS2017_6975}.

\section{Results and Discussions}
In this section, the performance of each module in the proposed teeth extraction approach is briefly addressed. Also, the proposed teeth extraction system is compared with other studies using different types of dental x-rays. Worth mentioning that as far as the accuracy is concerned, the ratio of correctly separated segments to the total number of parts is considered as the accuracy reported for teeth separation. Next, the results of caries detection with PaXNet is presented. Then, performance of the model on detecting caries in different stages is investigated. The contribution of each feature extractor network is also addressed by visual illustration of the proposed model's robustness.

\subsection{Experimental results}
For the teeth extraction task, the algorithm is applied on 42 Panoramic images, where the total number of teeth is 1229; 616 maxillary and 613 mandibular. Jaw region separation from the surrounding unwanted area is performed on the dataset, and the success rate of 40 out of 42 images is achieved. Hence, jaw extraction accuracy score is 95.23\%. It also failed to identify some of the wisdom teeth correctly. Thus, 4 maxillary and 2 mandibular wisdom teeth were also missed. The final number of remaining teeth is 582 in maxilla and 581 in mandible jaws.

Next, jaw separation is applied to 40 images comparing middle points and snake methods explained in the previous section. While middle points approach fails to separate the jaws correctly, and the crowns of one or more teeth are misclassified in many samples, snake approach demonstrates better performance and consistency. It also proves to work well even on closely-stacked-together jaws.

\begin{figure}[H]
    \begin{subfigure}{\linewidth}
      \centering
      \includegraphics[width=0.45\linewidth]{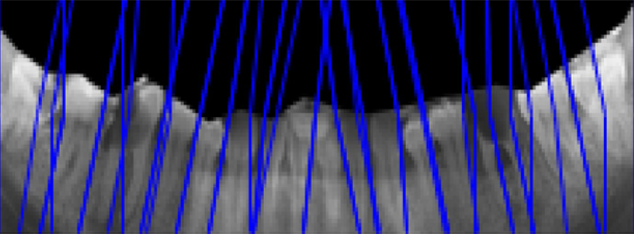}  
      \caption{}
      \label{fig:sub-first-ga}
    \end{subfigure}
    \hfill
    \begin{subfigure}{\linewidth}
      \centering
      \includegraphics[width=0.45\linewidth]{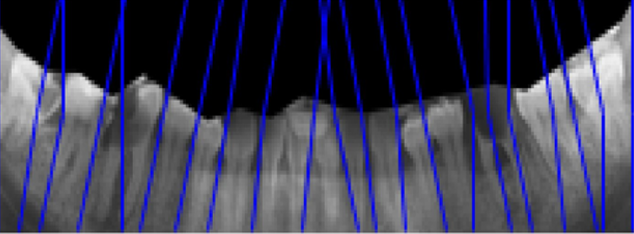}  
      \caption{}
      \label{fig:sub-second-ga}
    \end{subfigure}
    \hfill
    \begin{subfigure}{\linewidth}
      \centering
      \includegraphics[width=0.45\linewidth]{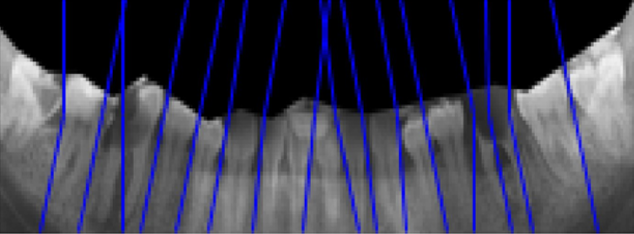}  
      \caption{}
      \label{fig:sub-third-ga}
    \end{subfigure}
    \caption{Tooth extraction procedure for a sample jaw: (a) preliminary genetic algorithm output (b) initial line removal output (c) final line removal output}
    \label{sample_isolation}
\end{figure}

After jaw separation, genetic algorithm is applied on extracted jaws, followed by line removal techniques to reduce the number of wrong lines, mostly passing through the teeth instead of teeth-gap valleys. Final results on maxillary and mandibular teeth, after applying line removal techniques, are presented in Table \ref{extraction_result}. Sample tooth isolation result after initial and final line removal steps is shown in Fig. \ref{sample_isolation}.

\begin{table}[H]
    \caption{Accuracy of tooth isolation in maxilla and mandible}
    \begin{center}
    \begin{tabular}{|c|c|c|c|}
    \hline
    \textbf{Jaw Type} & \textbf{Total Teeth} & \textbf{Isolated Teeth} & \textbf{Accuracy} \\ \hline
    Maxilla           & 582                  & 474                     & 81.44\%           \\ \hline
    Mandible          & 581                  & 428                     & 73.67\%           \\ \hline
    \end{tabular}
    \label{extraction_result}
    \end{center}
\end{table}

In the maxilla, the proposed algorithm works great isolating incisor teeth; however, molar and premolar teeth performance is mediocre. Most of the time, two central incisors are missed due to the lack of clarity in boundaries. This lack of transparency, like dark shadows, is typical for central incisors in Panoramic images, which is caused by the imaging method deficiency. To improve accuracy in mandibular teeth, model capability is increased by changing the cost function and line removal technique when dealing with mandibular images. Table \ref{extraction_comparison} briefly explains a comparison between our proposed teeth extraction method and other relevant research studies.

\begin{table}[H]
    \centering
    \caption{Comparison between proposed method and other teeth extraction research works}
    \label{extraction_comparison}
    \begin{tabular}{|c|c|c|c|c|}
    \hline
    
    \textbf{Tooth Extraction System} & \textbf{Image Type} & \textbf{Algorithm} & \textbf{Correct Upper} & \textbf{Correct Lower} \\ 
    
    \hline
    
    Abdel-Mottaleb et al. \cite{abdel2003challenges} & \multirow{4}{*}{Bitewing} & Integral Intensity Method & 169/195 - 85\% & 149/181 - 81\% \\ 
    
    \cline{1-1} \cline{3-5}
    Olberg and Goodwin \cite{olberg2016automated} & &
    Path-based Method & 300/336 - 89.3\% & 270/306 - 88.2\% \\
    
    \cline{1-1} \cline{3-5}
    Nomir et al. \cite{nomir2005system} &  & Integral Intensity Method & 329/391 - 84\% & 293/361 - 81\% \\ 
    
    \cline{1-1} \cline{3-5}
    Al-Sherif \cite{al2012new} &  & Energy-based Method & 1604/1833 - 87.5\% & 1422/1692 - 84\% \\ 
    
    \hline
    
    Ehsani Rad et al. \cite{rad2018automatic} & Periapical & Integral Intensity Method & \multicolumn{2}{c|}{Overall: 90.83\%} \\ 
    
    \hline
    
    Proposed Study & Panoramic & Genetic-based Method & 474/582 - 81.44\% & 428/581 - 73.67\% \\ 
    
    \hline
    \end{tabular}
\end{table}

Although we are applying teeth extraction on Panoramic images, which are noisy, include unwanted parts, and the structure boundaries of segments are ambiguous, acquired accuracy is in line with previous studies. Extracted teeth along with tooth images from the UESB dataset are labeled and used as the input to the proposed caries detection network.

PaXNet is trained using 319 samples with caries and 1519 healthy samples. In order to deal with class imbalance, the smaller class is re-sampled. Then, $80\%$ of data is used for training and $20\%$ is excluded as test set. Moreover, data augmentation is applied to the samples in the training process. Each sample in the dataset is rotated randomly and a random zoom and shift are applied as well, as listed in Table \ref{aug}.
 
\begin{table}[H]
    \caption{Image augmentation functions}
    \label{aug}
    \centering
    \begin{tabular}{|c|c|c|}
    \hline
    \textbf{Attribute} & \textbf{Parameter} & \textbf{Value}              \\ \hline
    Rotation           & Angle              & $0^{\circ}$ to $90^{\circ}$ \\ \hline
    Flip               & Axis               & Vertical and Horizontal     \\ \hline
    Brightness         & Scale              & 70\% to 130\%               \\ \hline
    Zoom               & Scale              & 90\% to 150\%               \\ \hline
    Width Shift        & Scale              & -20\% to 20\%               \\ \hline
    Height shift       & Scale              & -20\% to 20\%               \\ \hline
    \end{tabular}
\end{table}

\noindent A rotation range of $90^\circ$ covers the total possible rotation range with the help of horizontal and vertical flip. Darkening or brightening an image with a large scale will result in information loss, hence we adjust the brightness to mostly $30\%$ higher or lower than the raw input image. Zooming out of the image will help the model to see caries lesions with different scales, while zooming in can result in missing the caries of the image. Thus, we selected a zoom-in range of $10\%$ and a zoom-out range of $50\%$. The same rule of preventing the caries miss for images with positive label applies to the width/shift range. Since caries mostly happen in the edges of a tooth, shifting must be set to a small value, which is set to $20\%$ in this case. Worth mentioning that applying other augmentation methods, such as adding noise or shearing the image, whether resulted in worsening the accuracy or a negligible change in the model performance. Hence, these methods are excluded from the augmentation procedure.

\begin{figure}[H]
    \centerline{\includegraphics[width=0.3\linewidth ]{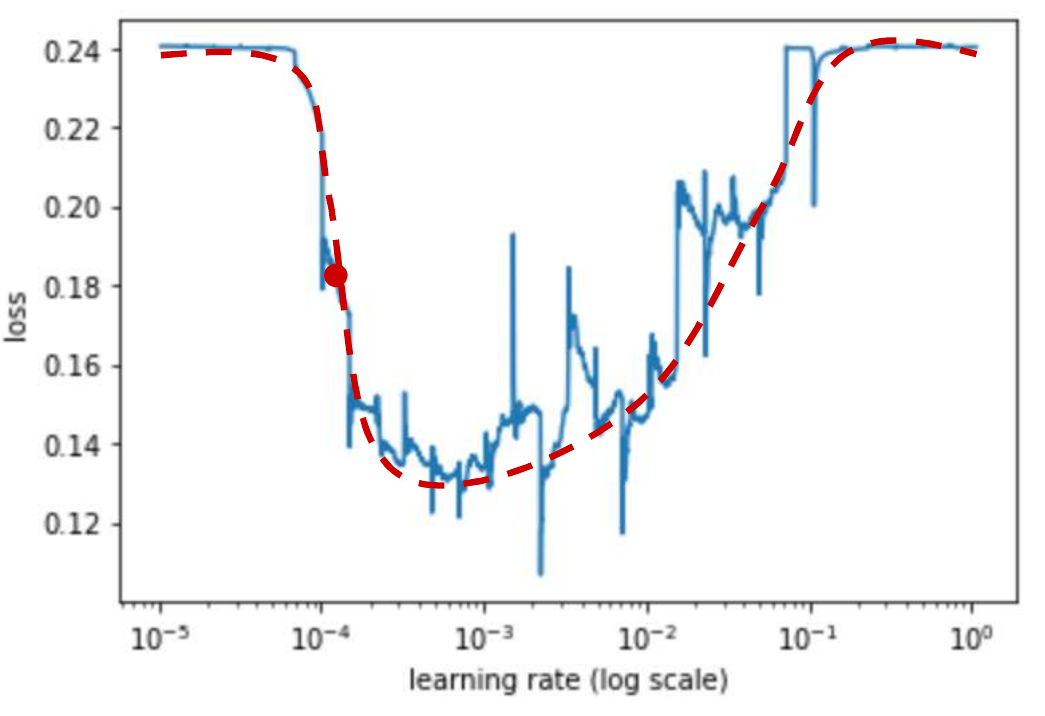}}
    \caption{loss vs learning rate and the optimal loss}
    \label{learning}
\end{figure}

Since CapsNet is very sensitive to learning rate, the optimal learning rate is calculated using the approach introduced in \cite{smith2017cyclical}. Fig. \ref{learning} illustrates the loss versus learning rate during 10 epochs of training while the learning is changing exponentially. 

According to Fig. \ref{learning} the learning rate is adjusted to $10^{-4}$. The model is trained for $850$ epochs with a batch size of $32$. After the training process, performance of the network is evaluated. Table \ref{acc} presents the results of the PaXNet over a dataset of 5948 extracted tooth images.

\begin{table}[H]
    \caption{Statistical performance of PaXNet}
    \label{acc}
    \centering
    \begin{tabular}{|c|c|c|c|c|c|}
    \hline
    \textbf{Dataset} & \textbf{Accuracy} & \textbf{Loss} & \textbf{Precision} & \textbf{Recall} & \textbf{F0.5-score}\\ \hline
    Training         & 91.23\%           & 0.13          & -                  & -               & -                  \\ \hline
    Test             & 86.05\%           & 0.15          & 89.41\%            & 50.67\%         & 0.78               \\ \hline
    \end{tabular}
\end{table}

\noindent While the accuracy score is high, considering the relatively smaller number of positive samples, the effect of outnumbering negative images must be decreased. Hence, f0.5-score is reported as the model result. Since carious teeth misclassified as healthy are more important than the false-positive cases, we should put more attention on minimizing false-positive ones. Thus, to increase the weight on precision and decrease the importance of recall, we selected f0.5-score as the best metric to measure the model performance.

\begin{figure}[H]
    \centering
    \includegraphics[width=0.5\linewidth]{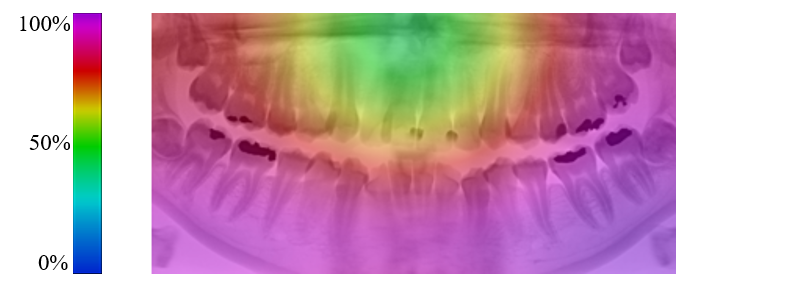}
    \caption{Location-based accuracy map of PaXNet. The average accuracy of the model in different parts of the jaw is plotted. Purple is interpreted as having the highest score, while blue is considered the lowest}
    \label{fig:loc_acc_map}
\end{figure}

To have a further look at how PaXNet is diagnosing caries based on the teeth location inside the jaw, a location-based accuracy map is drawn according to the accuracy of the model in the correct classification of each tooth category. To define a criterion, teeth are categorized into two classes that appear both in mandible and maxilla: molars-premolars and canines-incisors. As such, jaw is divided into $6$ regions. Fig. \ref{fig:loc_acc_map} shows the above-mentioned map.

\noindent Maxillary molars-premolars achieve the highest detection rate, while mandibular molars-premolars have a lower rate. Canines-incisors in both jaws have lower accuracy scores, resulting from a lack of sufficient carious teeth in the dataset. Worth mentioning that caries occurs in molars-premolars more than canines-incisors because of certain factors such as the salivary flow.

\subsection{Discussions}
Different stages for caries in the infected tooth can be considered. At first, the infected part is small. Then, the caries lesion grows, and a larger area is affected. The detection of severe cases has higher priority since they more likely need immediate treatment. To evaluate PaXNet performance in different tooth infection levels, samples with caries in the test set are divided into two categories, mild and severe. Then, the accuracy is computed for each group, as shown in Table \ref{stage}. 

\begin{table}[H]
    \caption{The accuracy of PaXNet for different infection stages}
    \label{stage}
    \centering
    \begin{tabular}{|c|c|}
    \hline
    \textbf{Category} & \textbf{Recall} \\ \hline
    Mild              & 69.44\%           \\ \hline
    Severe             & 90.52\%           \\ \hline
    Total             & 86.05\%           \\ \hline
    \end{tabular}
\end{table}

\noindent Severe decays usually appear as a larger demineralized area in tooth penetrating through enamel and dentine. In severe cases, it results in a total tooth collapse by destroying the pulpitis. Hence, As expected, the accuracy of the proposed model is notably higher in severe cases.

\begin{figure}[H]
    \centering
    \begin{subfigure}{0.25\linewidth}
      \centering
      \includegraphics[width=\linewidth]{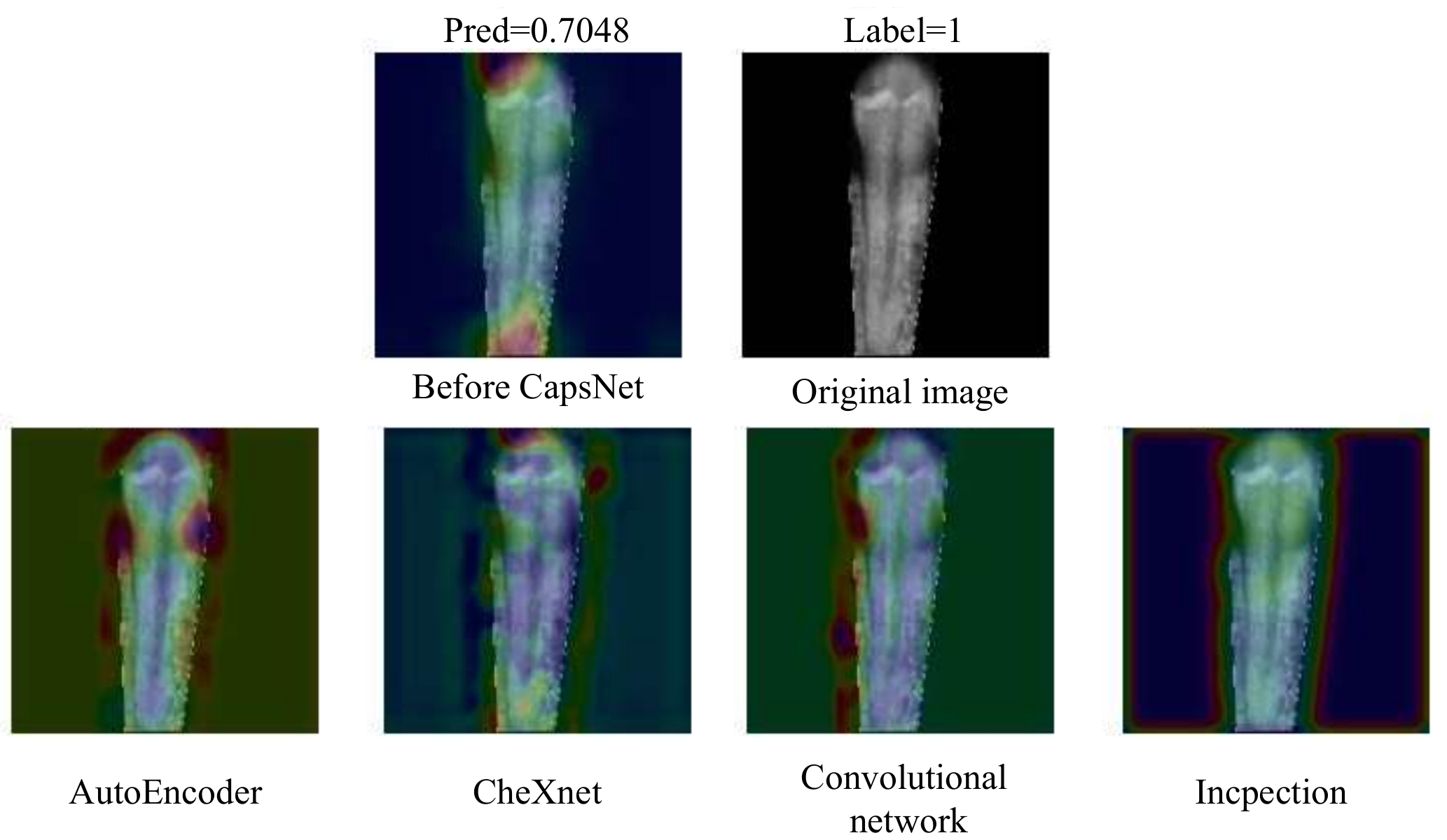}  
      \label{fig:sub-first}
    \end{subfigure}
    \begin{subfigure}{0.25\linewidth}
      \centering
      \includegraphics[width=\linewidth]{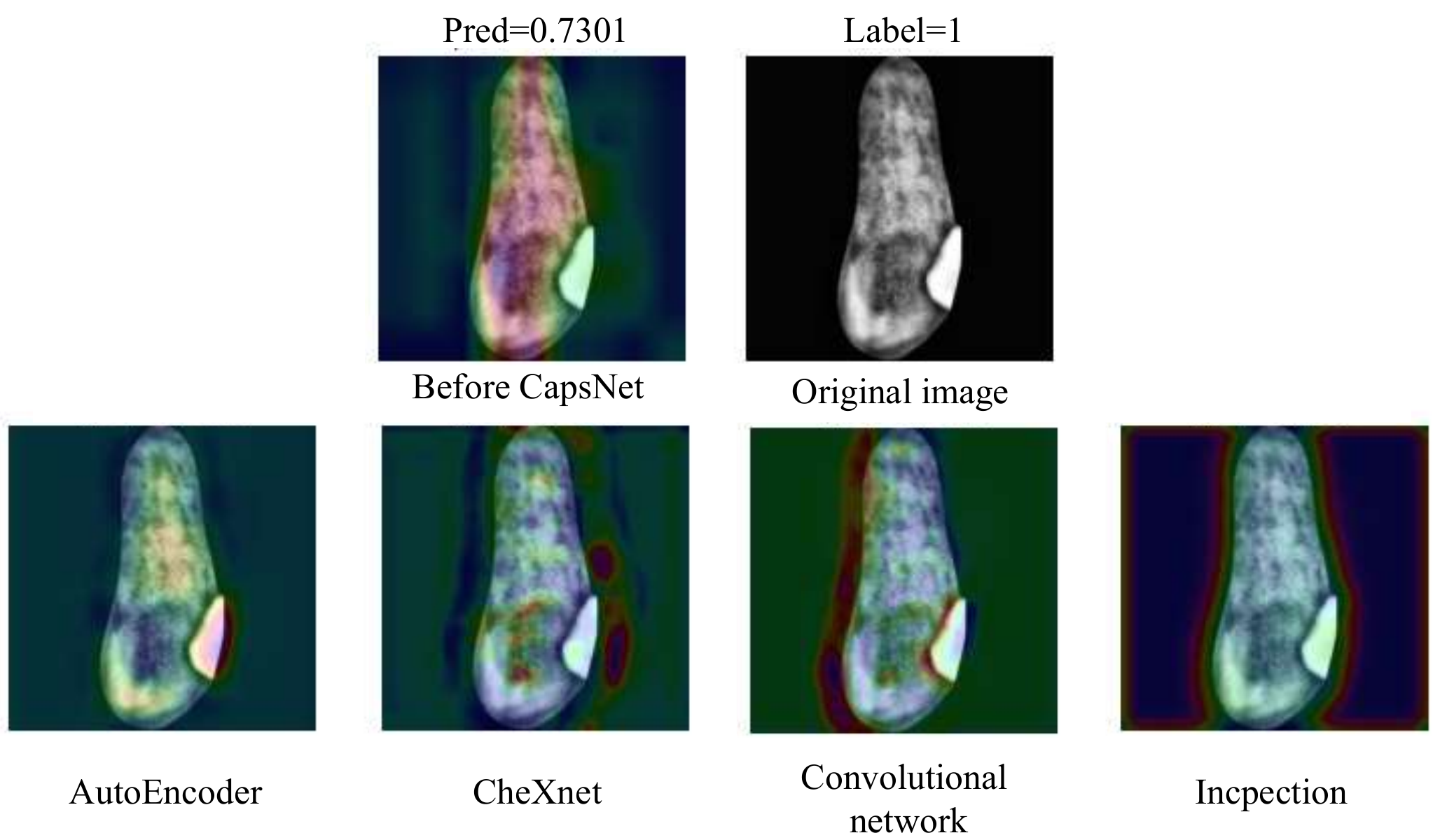}  
      \label{fig:sub-second}
    \end{subfigure}
    \begin{subfigure}{0.25\linewidth}
      \centering
      \includegraphics[width=\linewidth]{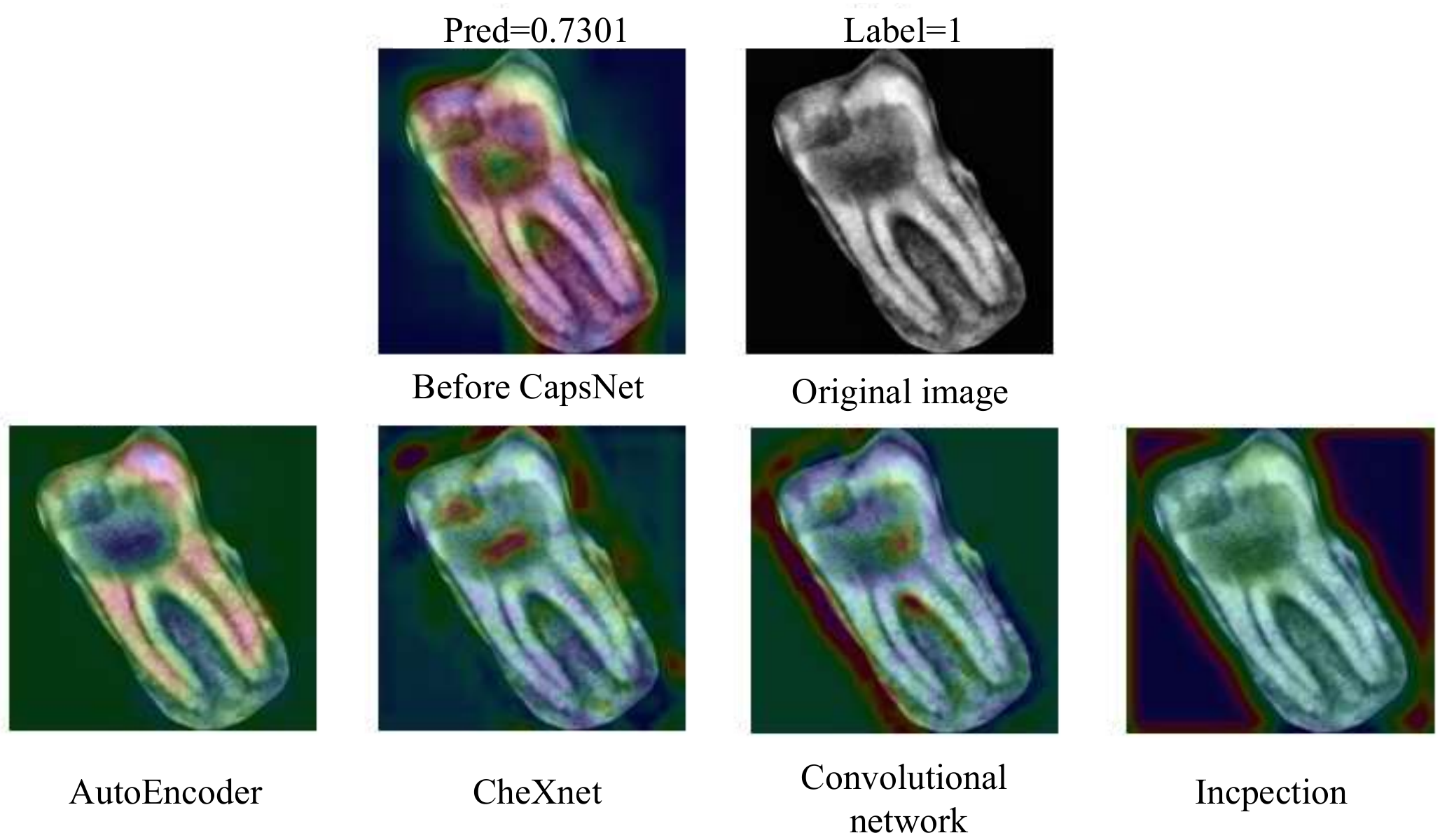}  
      \label{fig:sub-third}
    \end{subfigure}
    \begin{subfigure}{0.25\linewidth}
      \centering
      \includegraphics[width=\linewidth]{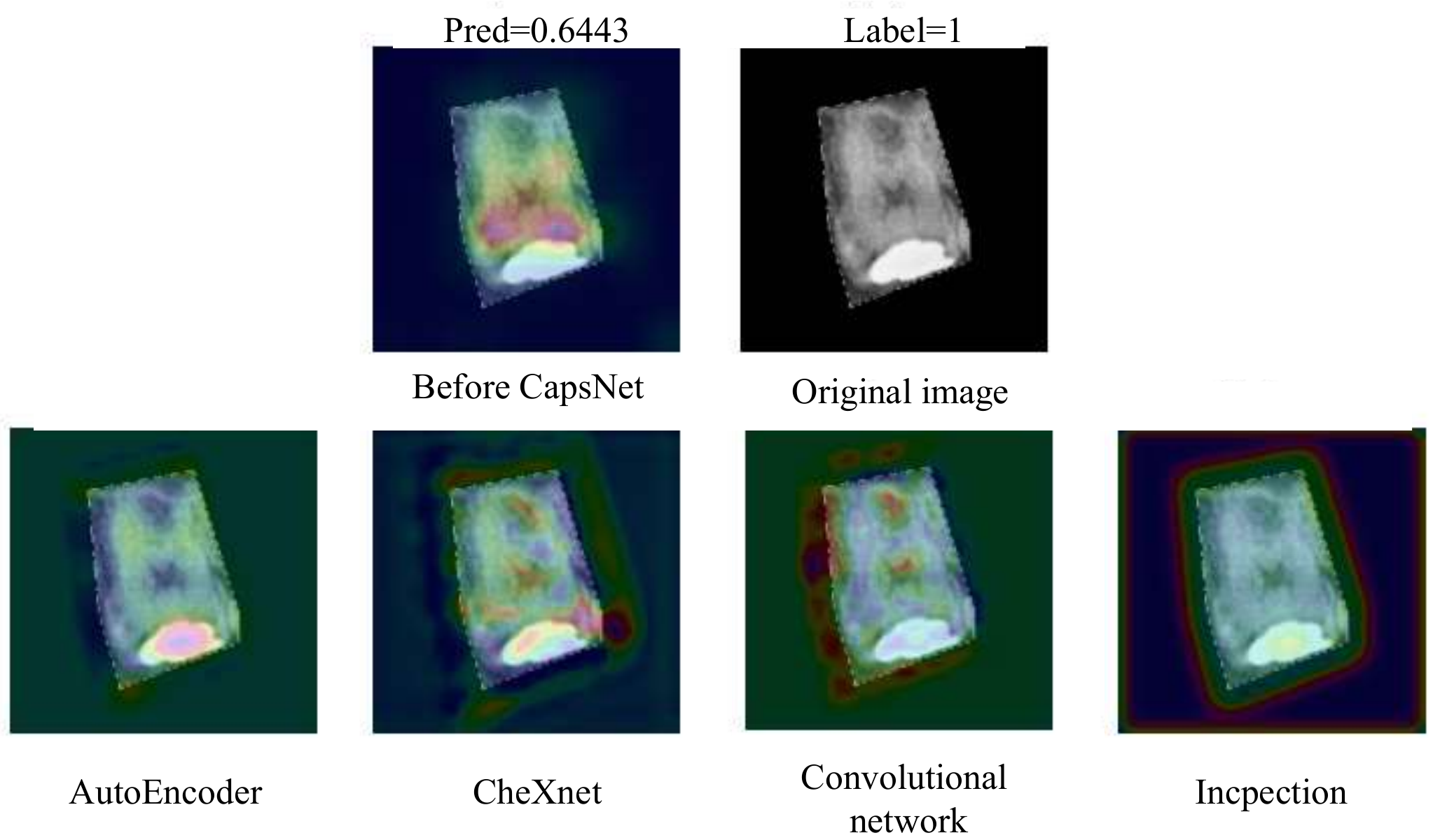}  
      \label{fig:sub-forth}
    \end{subfigure}
    \begin{subfigure}{0.25\linewidth}
      \centering
      \includegraphics[width=\linewidth]{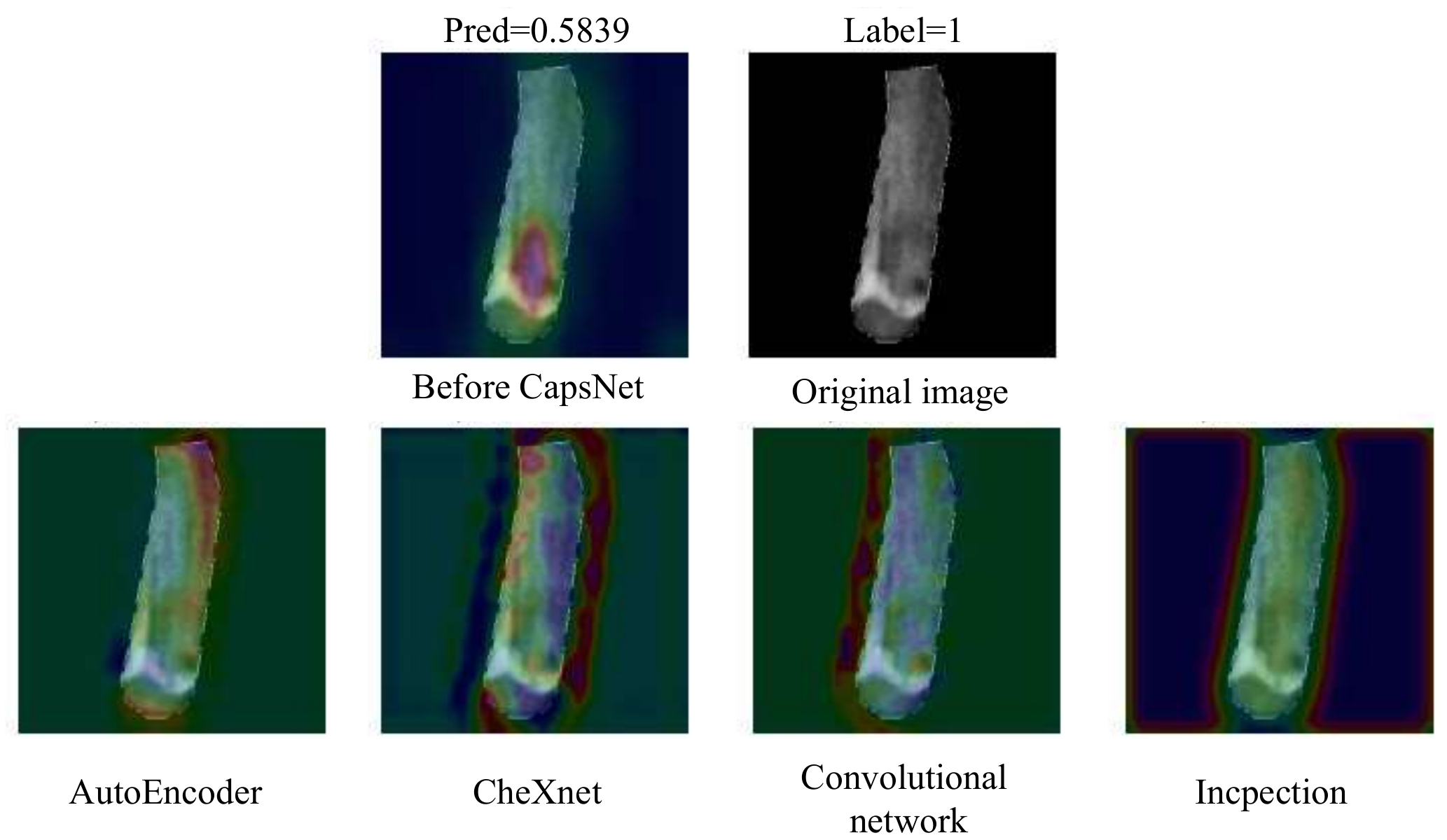}  
      \label{fig:sub-fifth}
    \end{subfigure}
    
    \caption{Five examples of Grad-CAM after each feature extraction network and before the capsule classifier}
    \label{grad_cam}
\end{figure}

The \textit{"right decision with wrong reason"} phenomenon can make the accuracy metrics distracting, especially when the dataset is relatively small. The computed accuracy can be a result of overfitting on this small dataset. Hence, the model might not perform this good on other samples. To make sure that the evaluated accuracy is a good reflection of the model's performance in facing new samples, the features contributing to the network's decision should be investigated. One of the most popular and effective approaches for feature visualization in CNN is Gradient-weighted Class Activation Mapping (Grad-CAM) \cite{selvaraju2017grad}. This method computed the heatmap regarding the location of the features most contributing to the final output using the gradient. The Grad-CAM of the last convolutional layer before the capsule network is plotted. Moreover, since these high-level features are the combination of the extracted features from four different models, the Grad-CAM of the convolutional layer before the concatenation is visualized as well. Fig. \ref{grad_cam} exhibits Grad-CAMs of five samples with caries.

\noindent The network is classifying these samples based on the true infected area in the tooth. Moreover, the larger the infected area is, the more confident the network becomes in caries detection. As far as feature extractors are concerned, each network is sensitive to a different type of feature. Table \ref{fe_acc} presents the test accuracy of PaXNet with different feature extraction units.

\begin{table}[H]
    \caption{The accuracy of PaXNet with different feature extractors}
    \label{fe_acc}
    \centering
    \begin{tabular}{|c|c|}
    \hline
    \textbf{Feature Extractor}           & \textbf{Test Accuracy} \\ \hline
    CNN                                  & 80.13\%               \\ \hline
    CNN-InceptionNet                     & 82.32\%               \\ \hline
    CNN-InceptionNet-CheXNet             & 84.67\%               \\ \hline
    CNN-InceptionNet-CheXNet-Autoencoder & 86.05\%               \\ \hline
    \end{tabular}
\end{table}

\noindent By combining these features, PaXNet detects caries based on the true infected area in the tooth. As a result, this model is capable of detecting caries correctly, and the reported accuracy is not the result of overfitting. Most importantly, in the face of new samples, a similar robust behaviour from the network is expected.

\begin{figure}[H]
    \centering
    \begin{subfigure}{0.7\linewidth}
        \centering
        \includegraphics[width=\linewidth]{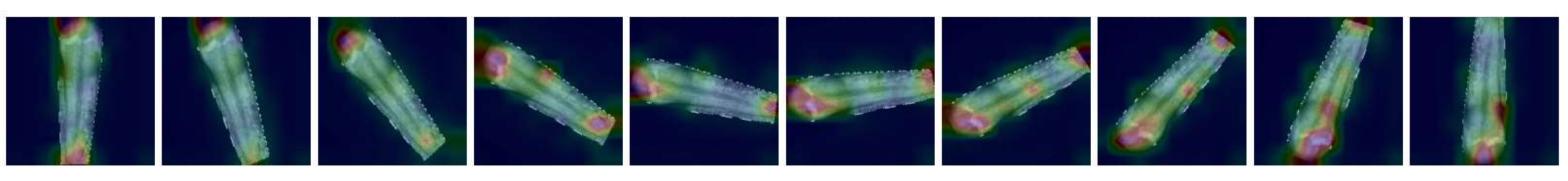}
        \caption{}
        \label{fig:v1.1}
    \end{subfigure}
    \hfill
    \begin{subfigure}{0.7\linewidth}
        \centering
        \includegraphics[width=0.7\linewidth]{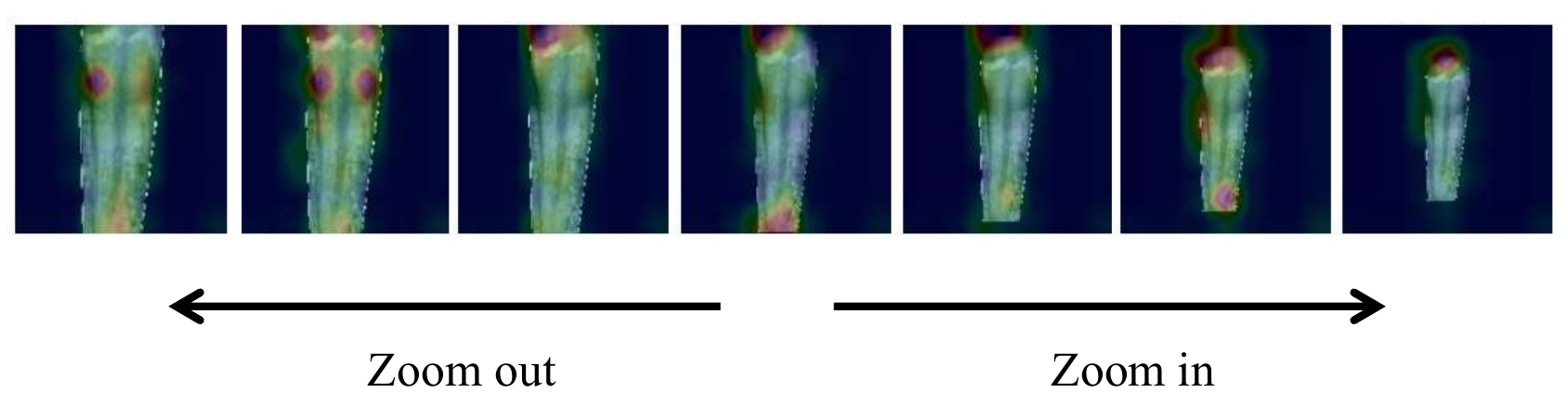}
        \caption{}
        \label{fig:v1.2}
    \end{subfigure}
    \begin{subfigure}{0.75\linewidth}
        \centering
        \includegraphics[width=0.5\linewidth]{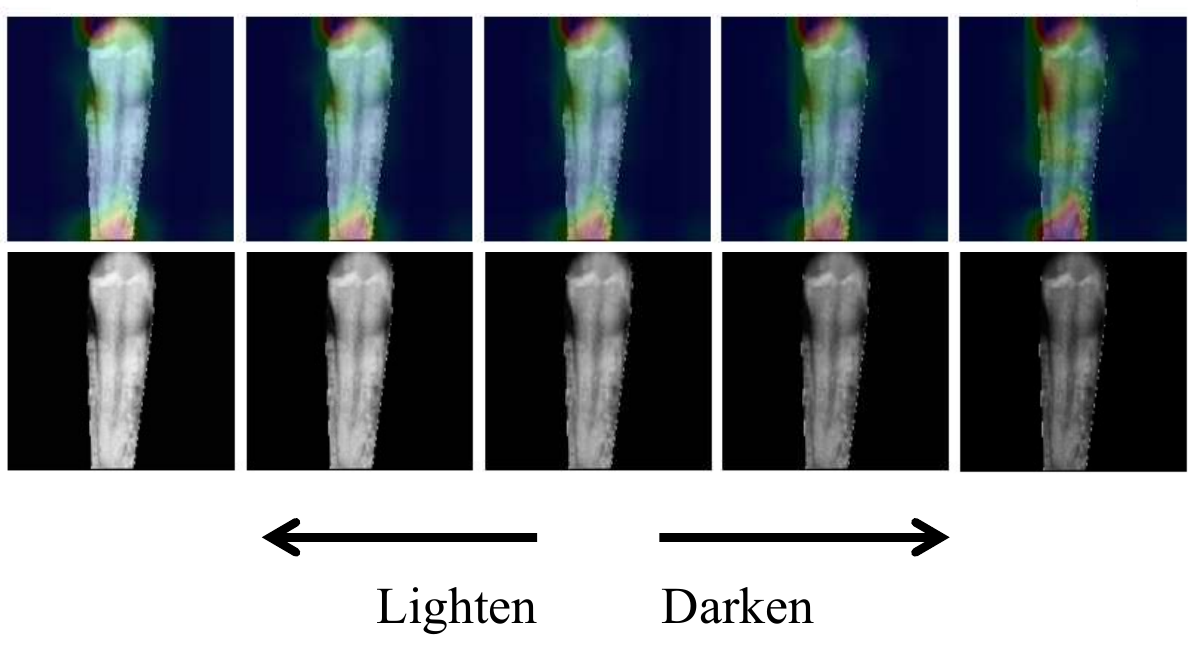}
        \caption{}
        \label{fig:v1.3}
    \end{subfigure}
    \hfill
\caption{The Grad-CAM visualization of an infected sample after transformation with (a) rotation, (b) zoom, and (c) brightness}
\label{test}
\end{figure}

The pose of a single tooth in x-ray images is typically vertical. However, there are some unusually posed teeth in some jaws. These problematic teeth are at higher risk of infection despite the smaller number of them in the dataset. In order to address this issue, data augmentation is performed in the training process. Fig. \ref{test} depicts the Grad-CAM of a sample with caries in various transformation.

\noindent The image is rotated from 0 to 180 degrees in 9 steps. Then, the image is scaled from $40\%$ to $160\%$. Finally, the mid-level brightness is altered $\pm 30\%$ using a quadratic function. Fig. \ref{test} illustrates that by applying various transformations, the proposed network is still classifying the sample accurately using correctly distinguished image features.

\section{Conclusion}
Dental decision support systems can help dentists by providing high-throughput diagnostic assistance to better detect dental diseases, such as caries, using radiographs. While capturing Panoramic dental radiography could be very helpful to see full patient dentition in a single image, detection of dental caries using dental Panoramic images is a very challenging task due to the low quality of the image and the ambiguity of decay regions. To help radiologists with dental caries diagnosis, we propose an automatic caries detection model using several feature extraction architectures and capsule network to predict labels from the extracted features. Grad-CAM visualization is used to validate our extracted features in terms of correct localization. Developed model could successfully hit an accuracy score of $86.05\%$ and an f0.5-score of $0.78$ on the test set. The achieved scores demonstrate model efficiency in concentrating on not missing carious teeth. Experimental results illustrate that the proposed combination of feature extraction modules achieves higher performance compared to a naïve CNN approach. Besides, PaXNet is also capable of categorizing caries lesions into two groups of mild ones with a recall of $69.44\%$ and severe ones with a recall score equal to $90.52\%$. Considering PaXNet as the first deep learning-based model to detect dental caries using Panoramic x-rays, achieved results are promising and acceptable, although not outstanding in comparison with previous works on image types with higher quality. To improve the results, two attributes are to be addressed: (1) increasing the number of Panoramic images, and specifically, expanding the number of carious teeth in the dataset. And (2) benefiting from a more sophisticated neural network, such as EfficientNet-based architectures, to boost the model's capability. Radiologist-provided precise annotations for caries regions will lead to accurate segmentation of caries lesions using U-Net-based segmentation models.

\newpage

\bibliography{main.bib}

\begin{thebibliography}{10}
\providecommand{\url}[1]{#1}
\csname url@samestyle\endcsname
\providecommand{\newblock}{\relax}
\providecommand{\bibinfo}[2]{#2}
\providecommand{\BIBentrySTDinterwordspacing}{\spaceskip=0pt\relax}
\providecommand{\BIBentryALTinterwordstretchfactor}{4}
\providecommand{\BIBentryALTinterwordspacing}{\spaceskip=\fontdimen2\font plus
\BIBentryALTinterwordstretchfactor\fontdimen3\font minus
  \fontdimen4\font\relax}
\providecommand{\BIBforeignlanguage}[2]{{%
\expandafter\ifx\csname l@#1\endcsname\relax
\typeout{** WARNING: IEEEtran.bst: No hyphenation pattern has been}%
\typeout{** loaded for the language `#1'. Using the pattern for}%
\typeout{** the default language instead.}%
\else
\language=\csname l@#1\endcsname
\fi
#2}}
\providecommand{\BIBdecl}{\relax}
\BIBdecl

\bibitem{selwitz2007dental}
R.~H. Selwitz, A.~I. Ismail, and N.~B. Pitts, ``Dental caries,'' \emph{The
  Lancet}, vol. 369, no. 9555, pp. 51--59, 2007.

\bibitem{amrollahi2016recent}
P.~Amrollahi, B.~Shah, A.~Seifi, and L.~Tayebi, ``Recent advancements in
  regenerative dentistry: A review,'' \emph{Materials Science and Engineering:
  C}, vol.~69, pp. 1383--1390, 2016.

\bibitem{beltran2005surveillance}
E.~D. Beltr{\'a}n-Aguilar, L.~K. Barker, M.~T. Canto, B.~A. Dye, B.~F. Gooch,
  S.~O. Griffin, J.~Hyman, F.~Jaramillo, A.~Kingman, R.~Nowjack-Raymer
  \emph{et~al.}, ``Surveillance for dental caries, dental sealants, tooth
  retention, edentulism, and enamel fluorosis; united states, 1988-1994 and
  1999-2002,'' 2005.

\bibitem{pitts2016white}
N.~Pitts and D.~Zero, ``White paper on dental caries prevention and
  management,'' \emph{FDI World Dental Federation}, 2016.

\bibitem{fejerskov2009dental}
O.~Fejerskov and E.~Kidd, \emph{Dental caries: the disease and its clinical
  management}.\hskip 1em plus 0.5em minus 0.4em\relax John Wiley \& Sons, 2009.

\bibitem{lira2009panoramic}
P.~H. Lira, G.~A. Giraldi, and L.~A. Neves, ``Panoramic dental x-ray image
  segmentation and feature extraction,'' in \emph{Proceedings of V workshop of
  computing vision, Sao Paulo, Brazil}, 2009.

\bibitem{tagliaferro2019caries}
E.~Tagliaferro, A.~V. Junior, F.~L. Rosell, S.~Silva, J.~L. Riley, G.~H.
  Gilbert, and V.~V. Gordan, ``Caries diagnosis in dental practices: Results
  from dentists in a brazilian community,'' \emph{Operative dentistry},
  vol.~44, no.~1, pp. E23--E31, 2019.

\bibitem{qu2011detection}
X.~Qu, G.~Li, Z.~Zhang, and X.~Ma, ``Detection accuracy of in vitro approximal
  caries by cone beam computed tomography images,'' \emph{European journal of
  radiology}, vol.~79, no.~2, pp. e24--e27, 2011.

\bibitem{akarslan2008comparison}
Z.~Akarslan, M.~Akdevelioglu, K.~Gungor, and H.~Erten, ``A comparison of the
  diagnostic accuracy of bitewing, periapical, unfiltered and filtered digital
  panoramic images for approximal caries detection in posterior teeth,''
  \emph{Dentomaxillofacial Radiology}, vol.~37, no.~8, pp. 458--463, 2008.

\bibitem{casamassimo1981radiographic}
P.~S. Casamassimo, ``Radiographic considerations for special
  patients-modifications, adjuncts, and alternatives,'' \emph{Ped Dent},
  vol.~3, no.~2, pp. 448--54, 1981.

\bibitem{underhill1988radiobiologic}
T.~E. Underhill, I.~Chilvarquer, K.~Kimura, R.~P. Langlais, W.~D. McDavid,
  J.~W. Preece, and G.~Barnwell, ``Radiobiologic risk estimation from dental
  radiology: Part i. absorbed doses to critical organs,'' \emph{Oral Surgery,
  Oral Medicine, Oral Pathology}, vol.~66, no.~1, pp. 111--120, 1988.

\bibitem{akkaya2006comparing}
N.~Akkaya, O.~Kansu, H.~Kansu, L.~Cagirankaya, and U.~Arslan, ``Comparing the
  accuracy of panoramic and intraoral radiography in the diagnosis of proximal
  caries,'' \emph{Dentomaxillofacial Radiology}, vol.~35, no.~3, pp. 170--174,
  2006.

\bibitem{naam2016algorithm}
J.~Naam, J.~Harlan, S.~Madenda, and E.~P. Wibowo, ``The algorithm of image edge
  detection on panoramic dental x-ray using multiple morphological gradient
  (mmg) method,'' \emph{International Journal on Advanced Science, Engineering
  and Information Technology}, vol.~6, no.~6, pp. 1012--1018, 2016.

\bibitem{jader2018deep}
G.~Jader, J.~Fontineli, M.~Ruiz, K.~Abdalla, M.~Pithon, and L.~Oliveira, ``Deep
  instance segmentation of teeth in panoramic x-ray images,'' in \emph{2018
  31st SIBGRAPI Conference on Graphics, Patterns and Images (SIBGRAPI)}.\hskip
  1em plus 0.5em minus 0.4em\relax IEEE, 2018, pp. 400--407.

\bibitem{lee2020application}
J.-H. Lee, S.-S. Han, Y.~H. Kim, C.~Lee, and I.~Kim, ``Application of a fully
  deep convolutional neural network to the automation of tooth segmentation on
  panoramic radiographs,'' \emph{Oral surgery, oral medicine, oral pathology
  and oral radiology}, vol. 129, no.~6, pp. 635--642, 2020.

\bibitem{haghanifar2020automated}
A.~Haghanifar, M.~M. Majdabadi, and S.-B. Ko, ``Automated teeth extraction from
  dental panoramic x-ray images using genetic algorithm,'' in \emph{2020 IEEE
  International Symposium on Circuits and Systems (ISCAS)}.\hskip 1em plus
  0.5em minus 0.4em\relax IEEE, 2020, pp. 1--5.

\bibitem{haghanifar2018dental}
A.~Haghanifar, A.~Amirkhani, and M.~R. Mosavi, ``Dental caries degree detection
  based on fuzzy cognitive maps and genetic algorithm,'' in \emph{Electrical
  Engineering (ICEE), Iranian Conference on}.\hskip 1em plus 0.5em minus
  0.4em\relax IEEE, 2018, pp. 976--981.

\bibitem{fried2020detecting}
D.~Fried, ``Detecting dental decay with infrared light,'' \emph{Optics and
  Photonics News}, vol.~31, no.~5, pp. 48--53, 2020.

\bibitem{casalegno2019caries}
F.~Casalegno, T.~Newton, R.~Daher, M.~Abdelaziz, A.~Lodi-Rizzini,
  F.~Sch{\"u}rmann, I.~Krejci, and H.~Markram, ``Caries detection with
  near-infrared transillumination using deep learning,'' \emph{Journal of
  dental research}, vol.~98, no.~11, pp. 1227--1233, 2019.

\bibitem{srivastava2017detection}
M.~M. Srivastava, P.~Kumar, L.~Pradhan, and S.~Varadarajan, ``Detection of
  tooth caries in bitewing radiographs using deep learning,'' \emph{arXiv
  preprint arXiv:1711.07312}, 2017.

\bibitem{choi2016boosting}
J.~Choi, H.~Eun, and C.~Kim, ``Boosting proximal dental caries detection via
  combination of variational methods and convolutional neural network,''
  \emph{Journal of Signal Processing Systems}, vol.~90, no.~1, pp. 87--97,
  2016.

\bibitem{lee2018detection}
J.-H. Lee, D.-H. Kim, S.-N. Jeong, and S.-H. Choi, ``Detection and diagnosis of
  dental caries using a deep learning-based convolutional neural network
  algorithm,'' \emph{Journal of dentistry}, vol.~77, pp. 106--111, 2018.

\bibitem{khan2020automated}
H.~A. Khan, M.~A. Haider, H.~A. Ansari, H.~Ishaq, A.~Kiyani, K.~Sohail,
  M.~Muhammad, and S.~A. Khurram, ``Automated feature detection in dental
  periapical radiographs by using deep learning,'' \emph{Oral Surgery, Oral
  Medicine, Oral Pathology and Oral Radiology}, 2020.

\bibitem{laishram2020detection}
A.~Laishram and K.~Thongam, ``Detection and classification of dental
  pathologies using faster-rcnn in orthopantomogram radiography image,'' in
  \emph{2020 7th International Conference on Signal Processing and Integrated
  Networks (SPIN)}.\hskip 1em plus 0.5em minus 0.4em\relax IEEE, 2020, pp.
  423--428.

\bibitem{rajpurkar2017chexnet}
P.~Rajpurkar, J.~Irvin, K.~Zhu, B.~Yang, H.~Mehta, T.~Duan, D.~Ding, A.~Bagul,
  C.~Langlotz, K.~Shpanskaya \emph{et~al.}, ``Chexnet: Radiologist-level
  pneumonia detection on chest x-rays with deep learning,'' \emph{arXiv
  preprint arXiv:1711.05225}, 2017.

\bibitem{silva2018automatic}
G.~Silva, L.~Oliveira, and M.~Pithon, ``Automatic segmenting teeth in x-ray
  images: Trends, a novel data set, benchmarking and future perspectives,''
  \emph{Expert Systems with Applications}, vol. 107, pp. 15--31, 2018.

\bibitem{martinez2011radiopacity}
F.~Mart{\'\i}nez-Rus, A.~M. Garc{\'\i}a, A.~H. de~Aza, and G.~Prad{\'\i}es,
  ``Radiopacity of zirconia-based all-ceramic crown systems.''
  \emph{International Journal of Prosthodontics}, vol.~24, no.~2, 2011.

\bibitem{goldberg2006genetic}
D.~E. Goldberg, \emph{Genetic algorithms}.\hskip 1em plus 0.5em minus
  0.4em\relax Pearson Education India, 2006.

\bibitem{sheta2012genetic}
A.~Sheta, M.~S. Braik, and S.~Aljahdali, ``Genetic algorithms: a tool for image
  segmentation,'' in \emph{2012 international conference on multimedia
  computing and systems}.\hskip 1em plus 0.5em minus 0.4em\relax IEEE, 2012,
  pp. 84--90.

\bibitem{NIPS2017_6975}
\BIBentryALTinterwordspacing
S.~Sabour, N.~Frosst, and G.~E. Hinton, ``Dynamic routing between capsules,''
  in \emph{Advances in Neural Information Processing Systems 30}, I.~Guyon,
  U.~V. Luxburg, S.~Bengio, H.~Wallach, R.~Fergus, S.~Vishwanathan, and
  R.~Garnett, Eds.\hskip 1em plus 0.5em minus 0.4em\relax Curran Associates,
  Inc., 2017, pp. 3856--3866. [Online]. Available:
  \url{http://papers.nips.cc/paper/6975-dynamic-routing-between-capsules.pdf}
\BIBentrySTDinterwordspacing

\bibitem{majdabadi2020msg}
M.~M. Majdabadi and S.-B. Ko, ``Msg-capsgan: Multi-scale gradient capsule gan
  for face super resolution,'' in \emph{2020 International Conference on
  Electronics, Information, and Communication (ICEIC)}.\hskip 1em plus 0.5em
  minus 0.4em\relax IEEE, 2020, pp. 1--3.

\bibitem{pal2018capsdemm}
A.~Pal, A.~Chaturvedi, U.~Garain, A.~Chandra, R.~Chatterjee, and S.~Senapati,
  ``Capsdemm: capsule network for detection of munro’s microabscess in skin
  biopsy images,'' in \emph{International Conference on Medical Image Computing
  and Computer-Assisted Intervention}.\hskip 1em plus 0.5em minus 0.4em\relax
  Springer, 2018, pp. 389--397.

\bibitem{tang2019capsurv}
B.~Tang, A.~Li, B.~Li, and M.~Wang, ``Capsurv: capsule network for survival
  analysis with whole slide pathological images,'' \emph{IEEE Access}, vol.~7,
  pp. 26\,022--26\,030, 2019.

\bibitem{iesmantas2018convolutional}
T.~Iesmantas and R.~Alzbutas, ``Convolutional capsule network for
  classification of breast cancer histology images,'' in \emph{International
  Conference Image Analysis and Recognition}.\hskip 1em plus 0.5em minus
  0.4em\relax Springer, 2018, pp. 853--860.

\bibitem{majdabadi2020capsule}
M.~M. Majdabadi and S.-B. Ko, ``Capsule gan for robust face super resolution,''
  \emph{Multimedia Tools and Applications}, pp. 1--14, 2020.

\bibitem{zhao2019deep}
T.~Zhao, Y.~Liu, G.~Huo, and X.~Zhu, ``A deep learning iris recognition method
  based on capsule network architecture,'' \emph{IEEE Access}, vol.~7, pp.
  49\,691--49\,701, 2019.

\bibitem{frosst2018darccc}
N.~Frosst, S.~Sabour, and G.~Hinton, ``Darccc: Detecting adversaries by
  reconstruction from class conditional capsules,'' \emph{arXiv preprint
  arXiv:1811.06969}, 2018.

\bibitem{szegedy2015going}
C.~Szegedy, W.~Liu, Y.~Jia, P.~Sermanet, S.~Reed, D.~Anguelov, D.~Erhan,
  V.~Vanhoucke, and A.~Rabinovich, ``Going deeper with convolutions,'' in
  \emph{Proceedings of the IEEE conference on computer vision and pattern
  recognition}, 2015, pp. 1--9.

\bibitem{ramachandran2017searching}
P.~Ramachandran, B.~Zoph, and Q.~V. Le, ``Searching for activation functions,''
  \emph{arXiv preprint arXiv:1710.05941}, 2017.

\bibitem{abdel2003challenges}
M.~Abdel-Mottaleb, O.~Nomir, D.~E. Nassar, G.~Fahmy, and H.~H. Ammar,
  ``Challenges of developing an automated dental identification system,'' in
  \emph{2003 46th Midwest Symposium on Circuits and Systems}, vol.~1.\hskip 1em
  plus 0.5em minus 0.4em\relax IEEE, 2003, pp. 411--414.

\bibitem{olberg2016automated}
J.-V. {\O}lberg and M.~Goodwin, ``Automated dental identification with lowest
  cost path-based teeth and jaw separation,'' \emph{Scandinavian Journal of
  Forensic Science}, vol.~22, no.~2, pp. 44--56, 2016.

\bibitem{nomir2005system}
O.~Nomir and M.~Abdel-Mottaleb, ``A system for human identification from x-ray
  dental radiographs,'' \emph{Pattern Recognition}, vol.~38, no.~8, pp.
  1295--1305, 2005.

\bibitem{al2012new}
N.~Al-Sherif, G.~Guo, and H.~H. Ammar, ``A new approach to teeth
  segmentation,'' in \emph{2012 IEEE International Symposium on
  Multimedia}.\hskip 1em plus 0.5em minus 0.4em\relax IEEE, 2012, pp. 145--148.

\bibitem{rad2018automatic}
A.~E. Rad, M.~S.~M. Rahim, H.~Kolivand, and A.~Norouzi, ``Automatic
  computer-aided caries detection from dental x-ray images using intelligent
  level set,'' \emph{Multimedia Tools and Applications}, vol.~77, no.~21, pp.
  28\,843--28\,862, 2018.

\bibitem{smith2017cyclical}
L.~N. Smith, ``Cyclical learning rates for training neural networks,'' in
  \emph{2017 IEEE Winter Conference on Applications of Computer Vision
  (WACV)}.\hskip 1em plus 0.5em minus 0.4em\relax IEEE, 2017, pp. 464--472.

\bibitem{selvaraju2017grad}
R.~R. Selvaraju, M.~Cogswell, A.~Das, R.~Vedantam, D.~Parikh, and D.~Batra,
  ``Grad-cam: Visual explanations from deep networks via gradient-based
  localization,'' in \emph{Proceedings of the IEEE international conference on
  computer vision}, 2017, pp. 618--626.

\end{thebibliography}
\bibliographystyle{IEEEtran}

\end{document}